\documentclass{article}

\PassOptionsToPackage{numbers,compress}{natbib}

\usepackage[main, final]{neurips_2026}

\usepackage[utf8]{inputenc}
\usepackage[T1]{fontenc}
\usepackage{hyperref}
\usepackage{url}
\usepackage{microtype}
\usepackage{xcolor}
\usepackage{graphicx}
\usepackage{subcaption}
\usepackage{booktabs}
\usepackage{multirow}
\usepackage{float}
\usepackage{algorithm}
\usepackage{algpseudocode}
\usepackage{amsmath}
\usepackage{amssymb}
\usepackage{amsfonts}
\usepackage{mathtools}
\usepackage{nicefrac}
\usepackage{amsthm}
\usepackage[capitalize,noabbrev]{cleveref}

\graphicspath{{./}}

\theoremstyle{plain}

\theoremstyle{definition}

\theoremstyle{remark}

\title{Beyond Masked Sparsity: SNACK Enables Truly Sparse Neural Networks on GPU}

\author{%
  Jafar Badour \\
  University of Twente \\
  \texttt{jafar.badour@utwente.nl}
  \And
  Maurice van Keulen \\
  University of Twente \\
  \texttt{m.vankeulen@utwente.nl}
  \And
  Elena Mocanu \\
  University of Twente \\
  \texttt{e.mocanu@utwente.nl}
}

\begin{document}
\maketitle


\begin{abstract}
Deep neural networks continue to grow in parameter count, driving up training and inference cost on GPUs. Sparse neural networks and Dynamic Sparse Training (DST) promise to reduce these costs, but most implementations rely on binary masks over dense tensors and recover little of the theoretical compute, memory, or energy savings. We propose SNACK, a truly sparse GPU layer that stores and computes only non-zero connections. SNACK exposes a simple PyTorch API for restructuring connections and backpropagating gradients entirely in the sparse paradigm, and ships SNACK-COO, a custom COO-format SpMM CUDA kernel with a batch-to-Streaming-Multiprocessor mapping tuned for the small-batch, high-sparsity regime typical of large-model training and single-stream inference. At the kernel level, SNACK is up to $7\times$ faster than the masked dense baseline (Dense+Mask) and competitive with cuSPARSE, Sputnik, and FlashSparse at 95\% sparsity. At 90\% sparsity, a single SNACK layer accelerates training by $8\times$ and $3.7\times$, and inference by $4\times$ and $2\times$, over Dense+Mask and fully dense layers, respectively, while using 72\% less memory than dense and substantially less energy. End-to-end, SNACK reduces GPT-2 peak training memory by up to 40\% and graph-style inference latency by $4.8\times$ over Dense+Mask at 99\% sparsity.
\end{abstract}

\vspace{-0.5em}
\section{Introduction}
\vspace{-0.5em}
Training and serving large neural networks is dominated by dense matrix multiplications whose cost scales with model size, raising real concerns about energy, hardware, and carbon emissions~\cite{strubell2019energy, schwartz2020green}. Large language models in particular routinely involve weight matrices of size around $10000\times 10000$, with both training and inference dominated by these multiplications. Sparse neural networks (SNNs) and Dynamic Sparse Training (DST)~\cite{mocanu2018scalable, evci2020rigging} aim to cut this cost by keeping most weights at zero throughout training, in principle reducing parameters and FLOPs quadratically. In practice, however, most SNN implementations simulate sparsity by applying a binary mask to a dense tensor~\cite{gale2019state, graesser2022state}: the dense matmul still runs underneath, and almost none of the theoretical compute, memory, or energy savings materialize on the GPU.
This leaves a gap between the promise of \textit{truly sparse} neural networks (where weights live in a sparse format end-to-end~\cite{pmlr-v139-liu21y}) and what is actually deployed. Mature high-performance SpMM(Sparse Matrix Multiplication) kernels such as cuSPARSE~\cite{naumov2010cusparse}, Sputnik~\cite{gale2020sparse}, and FlashSparse~\cite{shi2025flashsparse} cover several SpMM regimes well, but the very small batch sizes (down to batch~$=1$) typical of memory-constrained DST training and latency-bound single-stream inference on commodity GPUs remain underserved, so many fall back to a dense matmul times a binary mask, recovering little of the theoretical savings. We argue that this fallback \textit{should} and \textit{can} be avoided: with a small-batch, high-sparsity kernel, a truly sparse layer saves real time, memory, and energy over the masked-dense baseline. We further argue that for other regimes, current and future SpMM kernels can be plugged into our layer as backends, while still benefiting from its sparse gradient handling and sparse initialization.
\textbf{In this paper}, we close that gap with \textbf{SNACK} (\emph{Sparse Network Active CUDA Kernel}), which \textbf{contributes two pieces}. First, \textbf{SNACK-COO}: a custom COO-format SpMM CUDA kernel with a batch-to-Streaming-Multiprocessor mapping tuned for the small-batch, high-sparsity regime. Second, the \textbf{SNACK layer}: a truly sparse, backend-agnostic drop-in for \texttt{nn.Linear} with a sparse outer-product gradient and a memory-cheap sparse initializer, into which third-party SpMM kernels (cuSPARSE, Sputnik, FlashSparse) or our own SNACK-COO plug in as interchangeable backends. We demonstrate this second contribution end-to-end on a consumer-class RTX~3070 (6\,GB): dropping Sputnik into the SNACK layer (\texttt{SNACK(Sputnik)}) trains GPT-2 to validation perplexity statistically equivalent to \texttt{Dense+Mask} (the two backends cannot be seed-aligned because Sputnik's path through Algorithm~\ref{alg:uni_init} consumes a different RNG (Random Number Generator) stream from the dense Gaussian-then-mask path; \S\ref{sec:end_to_end_experiments}) while using up to $40\%$ less peak memory at $\eta=0.99$, so the same model trains and serves on a card that would otherwise be excluded by the dense weight-tensor memory bill.
On a single layer at $90\%$ sparsity (Section~\ref{sec:exps}), SNACK delivers:
\vspace{-0.5em}
\begin{itemize}
    \item vs.\ \textbf{Dense+Mask}: \textbf{$4\times$\,/\,$8\times$} faster inference/training, \textbf{$72\%$} less energy, \textbf{$73\%$} less memory.\vspace{-0.5em}
    \item vs.\ \textbf{Dense}: \textbf{$2\times$\,/\,$3.7\times$} faster inference/training, \textbf{$35\%$} less energy, \textbf{$72\%$} less memory.
\end{itemize}
\vspace{-0.5em}
End-to-end (Section~\ref{sec:end_to_end_experiments}), SNACK further cuts GPT-2 peak training memory by up to \textbf{$40\%$} and graph-style inference latency by \textbf{$4.8\times$} over Dense+Mask at $99\%$ sparsity. We anticipate that SNACK will serve as a starting point for further work on truly-sparse layers and per-regime SpMM kernels, both on the consumer-class accelerators where the memory bill is the binding constraint and on the datacenter-class hardware where the kernel-choice frontier is still evolving.
\vspace{-1em}
\section{Related Works}
\label{sec:related_works}
\vspace{-1em}
We organize prior work along three axes that bound our contribution: (i) sparsity-inducing training methods, (ii) GPU SpMM systems, and (iii) end-to-end truly sparse implementations. An extended discussion (covering early pruning, additional sparse-format details, and CPU/IPU truly-sparse work) is provided in Appendix~\ref{app:extended_related_works}.

\begin{table}[ht]
\centering
\footnotesize
\caption{DST-oriented methods that maintain a sparse representation throughout training. ``Truly Sparse'' means no dense weight tensor is materialized at any point; ``Unstructured'' means the method does not require a block, $N{:}M$, or other hardware-friendly sparsity pattern; ``Hardware-Aware'' means the method requires structured patterns or specialized accelerator support. SNACK is, to our knowledge, the first truly sparse PyTorch building block on commodity GPUs that supports unstructured sparsity end-to-end and exposes both forward and backward passes in the sparse paradigm.}
\label{tab:truly_sparse_methods}
\begin{tabular}{lccccc}
\toprule
Method & Truly Sparse & Unstructured & Hardware-Aware & Sparsity Phase & CPU/GPU/IPU \\
\midrule
Truly SNNs at Scale~\cite{curci2021truly} & \checkmark & \checkmark & & Training & CPU \\
SET-MLP~\cite{pmlr-v139-liu21y} & \checkmark & \checkmark & & Training & CPU \\
QuickSelection~\cite{Atashgahi2022} & \checkmark & \checkmark & & Training (Autoencoder) & CPU \\
Sparse-IFT~\cite{thangarasa2023sparse} & & & \checkmark & Training \& Inference & GPU \\
PopSparse~\cite{li2023popsparse} & \checkmark & & \checkmark & Training \& Inference & IPU \\
BLaST~\cite{okanovic2025blast} & & & \checkmark & Training \& Inference & GPU \\
\textbf{SNACK (ours)} & \checkmark & \checkmark & & Training \& Inference & GPU \\
\bottomrule
\end{tabular}
\end{table}
\vspace{-1em}
\paragraph{Sparsity-inducing training.} Sparsity in deep networks has been induced via post-hoc pruning~\cite{hassibi1993optimal, frankle2018lottery, frantar2023sparsegpt, zhu2017prune, liang2021pruning}, pruning-at-init~\cite{lee2019snip, tanaka2020pruning}, and Dynamic Sparse Training (DST)~\cite{bellec2018deep, mocanu2018scalable, dettmers2019sparse, evci2020rigging, mostafa2019parameter}, which adapts connectivity throughout training. With rare exceptions, all of these maintain a dense weight tensor with a binary mask~\cite{liu2020dynamic, gale2019state, graesser2022state}, recovering little of the theoretical compute or memory savings; \cite{yuan2021mest} attacks this from the optimizer side by storing optimizer states only for active parameters, which is complementary to the kernel-level fix we pursue. Hardware-friendly transformations such as Sparse-IFT~\cite{thangarasa2023sparse} and SPDF~\cite{thangarasa2023spdf} preserve dense FLOP counts to remain compatible with GEMM, which is orthogonal to the truly sparse direction we pursue. At LLM scale, \cite{agarwalla2024enabling} pretrains high-sparsity Llama variants for efficient deployment, and orthogonally, \cite{zhao2024galore} reduces optimizer-state memory via low-rank gradient projections; both target the same memory wall as SNACK but along different axes (sparse pretraining recipe and low-rank gradients, respectively, vs.\ a truly sparse weight tensor). End-to-end, these methods are typically validated on language modeling with GPT-2~\cite{radford2019gpt2} on WikiText-103~\cite{merity2016pointer} and similar benchmarks, which we adopt in Section~\ref{sec:end_to_end_experiments}. We discuss the broader literature in Appendix~\ref{app:extended_related_works}.
\vspace{-0.5em}
\paragraph{GPU SpMM systems.} cuSPARSE~\cite{naumov2010cusparse} and Sputnik~\cite{gale2020sparse} optimize SpMM for the moderate-sparsity, large-batch regime that dominates dense neural network training; Sputnik exceeds dense throughput starting at $\approx 71\%$ sparsity. FlashSparse~\cite{shi2025flashsparse} maps unstructured SpMM onto NVIDIA Tensor Cores, targeting graph workloads at moderate sparsity, e.g.\ GAMLP-style models~\cite{zhang2022graph} on the Open Graph Benchmark~\cite{hu2020open}. BLaST~\cite{okanovic2025blast} pairs a fused Triton block-SpMM kernel with a block-prune-and-grow recipe to sparsify Transformer MLPs up to $95\%$ on multi-GPU LLM pretraining, but requires structured (block) sparsity and keeps the dense weight tensor live during training. MOSparse~\cite{wang2025optimizing} addresses sparse-sparse multiplication (SpGEMM, a different operation than SpMM), and SparseRL~\cite{luo2026sparse} synthesizes SpMV kernels via reinforcement learning. Background on CSR/BSR/COO formats and earlier kernel-engineering work~\cite{bell2009implementing, merrill2016merge, yang2018design} is given in Appendix~\ref{app:extended_related_works}.
\vspace{-0.5em}
\paragraph{Truly sparse implementations and positioning.} Truly sparse training (no dense weight tensor at any point) has been demonstrated on CPU using SciPy/Cython data structures~\cite{pmlr-v139-liu21y, curci2021truly, Atashgahi2022} and on IPU hardware in PopSparse~\cite{li2023popsparse}, which relies on \emph{block} sparsity ($b\in\{4,8,16\}$) to align with vectorized IPU instructions. On commodity GPUs, hardware-supported $2{:}4$ structured sparsity (Ampere/Hopper Tensor Cores) caps usable sparsity at $50\%$ and requires re-pruning, so DST methods such as RigL~\cite{evci2020rigging} and SET~\cite{mocanu2018scalable}, which produce \emph{unstructured} masks, instead fall back to dense+mask. As summarized in Table~\ref{tab:truly_sparse_methods}, the combination of \emph{truly sparse, unstructured, commodity-GPU} with no hardware-specific sparsity support is, to our knowledge, unique to SNACK. Throughout the paper, ``sparsity'' refers to unstructured sparsity unless stated otherwise.

\vspace{-1.5em}
\section{Sparse Network Active CUDA Kernel (SNACK)}\label{sec:our_method}
\vspace{-0.5em}

\begin{figure}[H]
    \centering
    \includegraphics[width=0.55\linewidth]{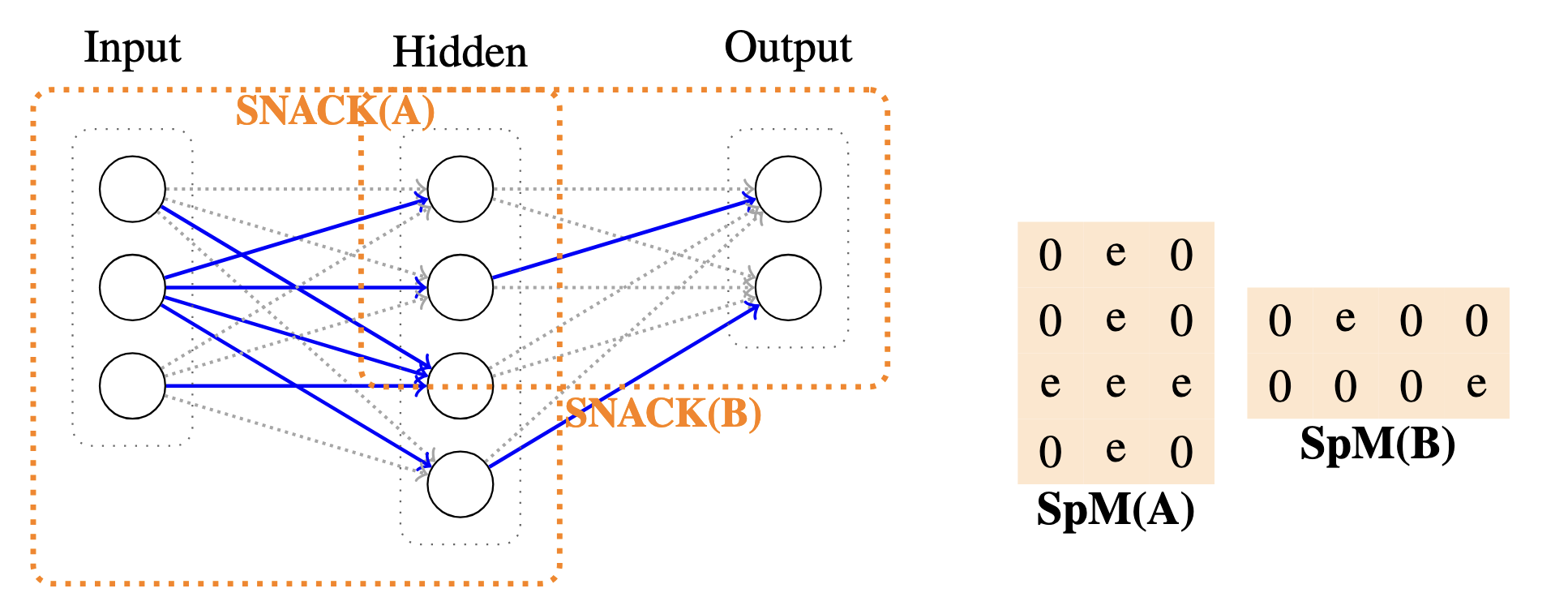}
    \caption{\textbf{SNACK layer overview.} Two stacked SNACK blocks ($A$ and $B$) act as drop-in replacements for standard \texttt{nn.Linear} layers. Each block stores its weight matrix $\mathrm{SpM}(\cdot)$ on the GPU as the COO triple $(\mathbf{A}_v, \mathbf{A}_r, \mathbf{A}_c)$ over its non-zeros only: the highlighted (blue) edges are the active sparse subnetwork that consumes memory and FLOPs, while the faded (grey) edges depict the underlying dense linear layer that SNACK \emph{never materializes} during training or inference. The forward pass is computed by SNACK-COO directly on the triple, the backward pass uses a sparse outer product over $\mathrm{supp}(W)$, and DST methods may add or drop active connections in place, so SNACK and dense blocks can be freely interleaved in a network without changing any caller code.}
    \label{fig:snack-overview}
\end{figure}

SNACK is a sparse-by-construction layer block (Figure~\ref{fig:snack-overview}) that drops in for \texttt{nn.Linear} and keeps weights in COO format end-to-end. The four subsections below cover the SpMM kernel (Sections~\ref{subsec:spmm} and \ref{subsec:coo_modified}), the uniform initializer (Algorithm~\ref{alg:uni_init}), sparse gradient computation (Section~\ref{subsec:sparse_grad}), and the PyTorch wrapping (Section~\ref{subsec:impl}). Appendix~\ref{app:snack_details} provides the full technical companion: a parallel-efficiency proof for SNACK-COO, a geometric-distribution analysis of the initializer's expected cost, the complete chain-rule and mini-batch gradient derivations, the CUDA/\texttt{autograd} implementation plumbing, and a summary of how end-to-end true sparsity is preserved across forward, backward, and topology updates.

\vspace{-2em}
\subsection{Sparse Matrix Multiplication}\label{subsec:spmm}
\vspace{-1em}

\noindent
\begin{minipage}[t]{0.51\textwidth}
\vspace{0pt}
We adopt a unified notation: $A \in \mathbb{R}^{m \times k}$ is the sparse weight matrix with sparsity $\eta$ (fraction of zero entries), $B \in \mathbb{R}^{k \times n}$ is the dense activation/input matrix where $n$ is the batch dimension, and $C = AB \in \mathbb{R}^{m \times n}$ is the output. Skipping zeros reduces the dense $\mathcal{O}(mkn)$ cost to $(1-\eta)\,mkn$, so runtime is proportional to the number of non-zeros $nnz = (1-\eta)\,mk$.

Algorithm~\ref{fig:sparse_mult} expresses this in the Coordinate (COO) format~\cite{bell2009implementing}, which stores each non-zero $A_{i,j}$ as a triple, i.e.\ as three arrays $(\mathbf{A}_v, \mathbf{A}_r, \mathbf{A}_c)$ over the 
\end{minipage}\hfill
\begin{minipage}[t]{0.46\textwidth}
\footnotesize
\vspace{-1pt}
\noindent\rule{\linewidth}{0.8pt}
\vspace{-1.5em}
\captionsetup{type=algorithm,labelfont=bf,textfont=normalfont}
\captionof{algorithm}{Sparse Matrix Multiplication}
\label{fig:sparse_mult}
\vspace{-0.5em}
\noindent\rule{\linewidth}{0.8pt}
\begin{flushleft}
\textbf{Input:} Sparse matrix $A \in R^{m \times k}$, activation matrix $B \in R^{k \times n}$ \\
\textbf{Output:} Matrix $C = A \times B$
\end{flushleft}
\begin{flushleft}
1. Initialize result matrix $C \in R^{m \times n}$ with all zeros \\
2. \textbf{for each} nonzero element $(i, j, A_z)$ in $A$ \textbf{do} \\
\hspace{1.5em} a. \textbf{for each} nonzero element $(j, l, B_z)$ in $B$ with matching $j$ \\
\hspace{3em} i. $C_{i, l} \gets C_{i, l} + A_z \cdot B_z$ \\
3. \textbf{return} $C$
\end{flushleft}
\vspace{-0.8em}
\noindent\rule{\linewidth}{0.8pt}
\end{minipage}
$nnz$ non-zeros (full definition in Appendix~\ref{app:sparse_ops}); CSR or BSR are interchangeable. Standard COO implementations distribute rows of $A$ across CUDA cores and treat $n$ as an outer loop, which underutilizes SMs at small $n$ and motivates Section~\ref{subsec:coo_modified}.
\subsection{SNACK-COO: Batch-to-SM Mapping}\label{subsec:coo_modified}

The standard COO scheme assigns one row of $A$ per CUDA thread and treats the batch dimension as an outer loop, yielding $O(nm)$ thread tasks and leaving SMs idle whenever $n \ll m$. Our proposed SNACK-COO instead launches a 2D block grid $(N_b, n)$ with $N_b = \lceil nnz / T \rceil$ for a thread-block size $T$ (typically $256$): block $(i, b)$ owns the $i$-th chunk of $T$ non-zeros and computes its contribution to output column $b$. Thread $\tau$ in block $(i, b)$ reads the COO triple at index $k = iT + \tau$ and atomically accumulates $C_{A_r[k],\,b} \mathrel{+}= A_v[k]\cdot B_{A_c[k],\,b}$. The task pool becomes $N_b \times n$ blocks (i.e.\ $nnz \times n$ thread-tasks), so SM utilization tracks the non-zero workload rather than the matrix height, which is decisive at small $n$. Appendix~\ref{app:snack_details} formalizes this with the parallel-efficiency inequality $\epsilon_p^{\text{SNACK-COO}} > \epsilon_p^{\text{standard}}$ for $n \ll m$, and shows why batch-wise blocks reduce atomic contention compared to row-wise distribution.
\vspace{-1em}
\subsection{Uniform Connection Initializer}
\vspace{-1em}
Most DST methods initialize the mask by drawing an $m \times k$ tensor of random numbers and thresholding, which materializes a dense intermediate and can OOM for large layers. Algorithm~\ref{alg:uni_init} avoids this: it samples $nnz = (1-\eta) m k$ unique index pairs uniformly at random and never builds a dense matrix. The expected cost is $\mathcal{O}(\zeta \cdot nnz)$, where $\zeta$ is the expected number of trials before drawing a fresh pair.

\noindent
\begin{minipage}[t]{0.45\textwidth}
At step $i$, the probability of drawing a fresh pair is $p_i = 1 - i/(mk)$, so by the geometric distribution the expected number of trials at that step is $1/p_i = mk/(mk - i)$. Summing over the $nnz$ unique pairs and applying the harmonic approximation $\sum_{i=a}^{b} 1/i \approx \ln(b/a)$ yields $\sum_{i=0}^{nnz} \frac{mk}{mk-i} \approx -mk \ln \eta$ for $\eta \in (0, 1)$, i.e.\ essentially linear in $nnz$. Collisions dominate as $\eta \to 0$ (dense regime) but vanish as $\eta \to 1$, so the rejection penalty $\zeta = -\ln \eta / (1-\eta)$ is bounded and small in the high-sparsity regime DST targets. The full derivation is in Appendix~\ref{app:snack_details}.
\end{minipage}\hfill
\begin{minipage}[t]{0.5\textwidth}
\vspace{-4pt}
\footnotesize
\vspace{-1pt}
\noindent\rule{\linewidth}{0.8pt}
\vspace{-1.5em}
\captionsetup{type=algorithm,labelfont=bf,textfont=normalfont}
\captionof{algorithm}{Uniform Initializer}
\label{alg:uni_init}
\vspace{-0.5em}
\noindent\rule{\linewidth}{0.8pt}
\begin{flushleft}
\textbf{Input:} Sparse matrix of size $m \times k$, target sparsity $\eta$ \\
\textbf{Output:} Set $\mathcal{C} = \{c_t = \langle i, j \rangle\}$ of $(1-\eta) \cdot m \cdot k$ unique indices
\end{flushleft}
\begin{flushleft}
1. Initialize empty index set $\mathcal{C}$, hash set $S$ \\
2. \textbf{for} $t = 1$ to $(1-\eta) \cdot m \cdot k$ \textbf{do} \\
\hspace{1.5em} a. \textbf{repeat} \\
\hspace{3em} i. $i \leftarrow \texttt{getrand}(m) + 1$ \\
\hspace{3em} ii. $j \leftarrow \texttt{getrand}(k) + 1$ \\
\hspace{3em} iii. \textbf{if} $(i, j) \in S$ \textbf{then} continue \\
\hspace{3em} iv. \textbf{else} add $(i, j)$ to $S$ \\
\hspace{5em} $c_t \leftarrow \langle i, j \rangle$ \\
\hspace{5em} \textbf{break} \\
\hspace{1.5em} b. \textbf{until} a new pair is added \\
3. \textbf{return} $\mathcal{C}$
\end{flushleft}
\vspace{-0.5em}
\noindent\rule{\linewidth}{0.8pt}
\end{minipage}
\vspace{-3em}
\subsection{Sparse Gradient Computation}\label{subsec:sparse_grad}
\vspace{-1em}
For a SNACK layer with sparse weight $W \in \mathbb{R}^{m \times n}$, input $X \in \mathbb{R}^{n \times b}$, and pre-activation $A = WX$, the chain rule yields $\frac{\partial \mathcal{L}}{\partial W} = (\frac{\partial \mathcal{L}}{\partial A})\,X^\top$ and $\frac{\partial \mathcal{L}}{\partial X} = W^\top (\frac{\partial \mathcal{L}}{\partial A})$. The input-gradient term is sparse-times-dense and reuses our SpMM directly since $W^\top$ inherits $W$'s sparsity. The weight-gradient term is the cost driver: a naive outer product produces a dense $m \times n$ matrix the size of the unpruned $W$. SNACK instead computes $(\frac{\partial \mathcal{L}}{\partial W})_{i,j} = \sum_{t=1}^{b} \nabla Y_{i,t}\,X_{j,t}$ \emph{only at indices $(i,j) \in \mathrm{supp}(W)$}, dropping the cost from $\mathcal{O}(mn b)$ to $\mathcal{O}(nnz \cdot b)$ while preserving SGD correctness (frozen entries see zero gradient anyway). Activation functions operate element-wise and are agnostic to $W$'s sparsity~\cite{hagan1997neural}. Appendix~\ref{app:snack_details} gives the complete derivation: the chain rule for $\partial \mathcal{L}/\partial W$ and $\partial \mathcal{L}/\partial X$, the entry-wise expansion that motivates restricting work to $\mathrm{supp}(W)$, and the generalization to mini-batches as a sum of $b$ rank-1 sparse outer products.

\noindent
\begin{minipage}[t]{0.52\textwidth}
\vspace{0pt}
The matrix on the right is $\nabla Y \otimes X = (\nabla Y)\,X^\top$, the entry-wise expansion of the outer product. Dense backprop fills all $mn$ entries; SNACK computes only those at $(i,j) \in \mathrm{supp}(W)$. For a single non-zero $\langle 2,1 \rangle$, the boxed cell $X_2 \nabla Y_1$ is the only entry materialized. Across forward, gradient, and topology updates, we maintain end-to-end true sparsity: no step ever instantiates a dense $m \times n$ tensor.
\end{minipage}\hfill
\begin{minipage}[t]{0.46\textwidth}
\vspace{0pt}
\noindent\rule{\linewidth}{0.8pt}
\vspace{0.15em}
\[
(\nabla Y)\,X^\top =
\begin{bmatrix}
X_{1} \nabla Y_{1} & \dots & X_{1} \nabla Y_{n} \\
\boxed{X_{2} \nabla Y_{1}} & \dots & X_{2} \nabla Y_{n} \\
\vdots & \vdots & \vdots \\
X_{m} \nabla Y_{1} & \dots & X_{m} \nabla Y_{n}
\end{bmatrix}
\]
\vspace{-0.25em}
Boxed entry computed iff $\langle 2,1 \rangle \in \mathrm{supp}(W)$.
\vspace{0.2em}
\noindent\rule{\linewidth}{0.8pt}
\end{minipage}

\subsection{Implementation}\label{subsec:impl}
\vspace{-0.5em}
SNACK is exposed as a PyTorch \texttt{nn.Module} that drops in for \texttt{nn.Linear} (Figure~\ref{fig:snack-overview}): forward and backward are wired through a custom \texttt{torch.autograd.Function} backed by the SNACK-COO CUDA kernel and a sparse outer-product CUDA kernel for $\frac{\partial \mathcal{L}}{\partial W}$. Weights live entirely on the GPU as the COO triple $(\mathbf{A}_v, \mathbf{A}_r, \mathbf{A}_c)$ and are never materialized as a dense $m \times n$ tensor at any point: initialization, forward, backward, or topology updates. The SpMM backend is swappable: SNACK-COO is the default for the small-batch DST regime, but any third-party SpMM (Sputnik, FlashSparse, cuSPARSE) can be plugged in for regimes where it is faster, while still inheriting SNACK's sparse outer-product gradient and uniform initializer. Wrapper, kernel, and \texttt{autograd} plumbing details are in Appendix~\ref{app:snack_details}; code is available at \url{https://github.com/JafarBadour/Snack-SNN}.

\vspace{-1em}
\section{Evaluation and Results}
\label{sec:exps}
\vspace{-0.5em}

We evaluate SNACK at three increasingly demanding levels: (i) the SpMM kernel in isolation (Section~\ref{subsec:exp_kernel}), (ii) a single sparse layer in a $100$-iteration MLP-style training cycle (Section~\ref{subsec:exp_layer}), and (iii) end-to-end on GPT-2 and a graph workload (Section~\ref{sec:end_to_end_experiments}). Unless stated otherwise, all micro-benchmarks run on an RTX~3070 (6\,GB) at sparsities $\eta\in\{0.9, 0.95, 0.99\}$ and small leading dimensions, the regime SNACK-COO targets. We report CUDA time, peak GPU memory, energy, and parameter count; per-input-size sweeps and additional figures are deferred to Appendix~\ref{app:perf_bench}.

\vspace{-0.5em}
\subsection{SpMM Kernel: SNACK-COO vs.\ State-of-the-Art}\label{subsec:exp_kernel}
\vspace{-0.5em}

\begin{figure}[t]
    \centering
\includegraphics[width=0.98\linewidth]{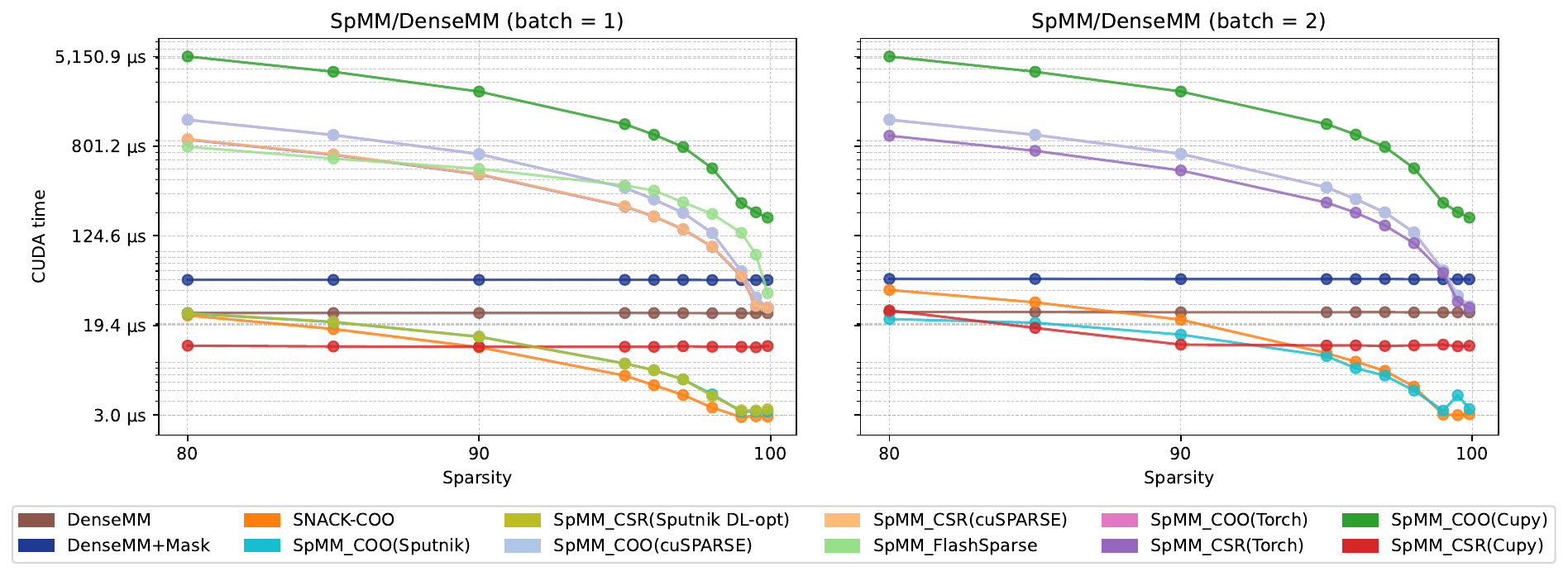}
    \caption{\textbf{SpMM CUDA time vs.\ sparsity at fixed batch sizes $\{1,2\}$ on a $5000\times 5000$ matrix.} Each panel sweeps sparsity for one batch size. SNACK-COO (ours, orange) stays below the Dense+Mask baseline across these small-batch settings and remains competitive with the dense reference.}
    \label{fig:sparse-methods-comp-p1}
\end{figure}

We benchmark SNACK-COO against eleven SpMM implementations~\cite{frostig2018compiling, jax2018github, nishino2017cupy, 10.5555/3454287.3455008} from PyTorch (CSR/COO), CuPy (CSR/COO), JAX (BSR), cuSPARSE (CSR/COO), Sputnik (CSR/COO), and FlashSparse on a $5000\times 5000$ sparse matrix at $95\%$ sparsity, against a $(\text{batch}\times 5000)$ dense matrix, averaged over $1000$ multiplications. Table~\ref{tab:sparse_performance} reports CUDA time at batch~$1$ across four matrix sizes: Our SNACK-COO is the fastest sparse method on every leading-dimension size we tested and the only sparse implementation that beats Dense+Mask, achieving a $7\times$ speedup over Dense+Mask and a $3.5\times$ speedup over the dense baseline at $5000\times 5000$. Figure~\ref{fig:sparse-methods-comp-p1} reports sparsity sweeps at fixed batch sizes $\{1,2\}$ and shows the same advantage across the small-batch regime. Appendix~\ref{app:a100_spmm} replicates this sweep on a datacenter-class NVIDIA~A100 across six leading-dimension sizes ($500$--$10000$) at batch~$\{1, 4\}$: at batch~$1$ SNACK-COO retains the per-column lead at every size ($7.5\times$ over Dense+Mask, $3.7\times$ over dense at $5000{\times}5000$), with \texttt{SpMM\_CSR(Cupy)} closing to within $13\%$ only at $\{8500, 10000\}$; at batch~$4$ SNACK-COO leads at $\{500, 1000\}$ and is overtaken from $5000{\times}5000$ upward by the two Sputnik variants (within ${\pm}10\%$ of each other at every size), the regime crossover predicted by the parallel-efficiency analysis (Appendix~\ref{app:snack_details}) and motivating the swappable backend of Section~\ref{subsec:impl}.

\begin{table*}[t]
\footnotesize
\centering
\caption{Performance comparison of sparse matrix multiplication implementations. SpM sparsity is 95\%, batch size of 1. Our SNACK-COO method outperforms other methods (7x reduction in time vs DenseMM+Mask).  $^\dagger$Simulates sparsity.}
\label{tab:sparse_performance}
\begin{tabular}{lccccc}
\toprule
\textbf{Method} & is sparse & \multicolumn{4}{c}{CUDA (ms)}   \\
\cmidrule(lr){3-6}
& & 500x500 & 1000x1000 & 5000x5000 & 10000x10000 \\
\midrule
SpMM\_CSR(Torch) & \multirow{12}{*}{Yes} & 27.05 & 26.84 & 231.83 & 885.33 \\
SpMM\_CSR(Cupy) &   & 11.95 & 11.66 & 12.27 & 17.90 \\
SpMM\_COO(Torch) &   & 27.41 & 29.10 & 341.60 & 1396.26 \\
SNACK-COO (ours) &  & \textbf{2.43} & \textbf{2.38} & \textbf{7.28} & \textbf{15.17} \\
SpMM\_COO(Cupy) &  & 207.86 & 215.45 & 1231.38 & 4946.64\\
SpMM\_BSR(Jax) &  & 88.68 & 75.78 & 126.39 & 320.32 \\
SpMM\_CSR(cuSPARSE) &  & 27.14 & 27.47 & 225.09 & 858.96 \\
SpMM\_COO(cuSPARSE) &  & 26.56 & 29.15 & 333.83 & 1362.39 \\
SpMM\_CSR(Sputnik DL-opt) &  & 3.21 & 3.31 & 8.93 & 25.37 \\
SpMM\_COO(Sputnik) &  & 3.22 & 3.22 & 8.89 & 25.36 \\
FlashSparse &  & 29.85 & 39.11 & 360.35 & 1180.53 \\
DenseMM+Mask & $^\dagger$ & 4.32 & 4.24 & 50.46 & 193.91 \\
\midrule
DenseMM(baseline) & no &\textbf{2.17 }& \textbf{2.13} & 25.36 & 97.44 \\
\bottomrule
        \end{tabular}
\end{table*}

\vspace{-0.5em}
\subsection{Single-Layer Training Cycle}\label{subsec:exp_layer}
\vspace{-0.5em}

\begin{figure}[t]
    \centering
    \includegraphics[width=\linewidth]{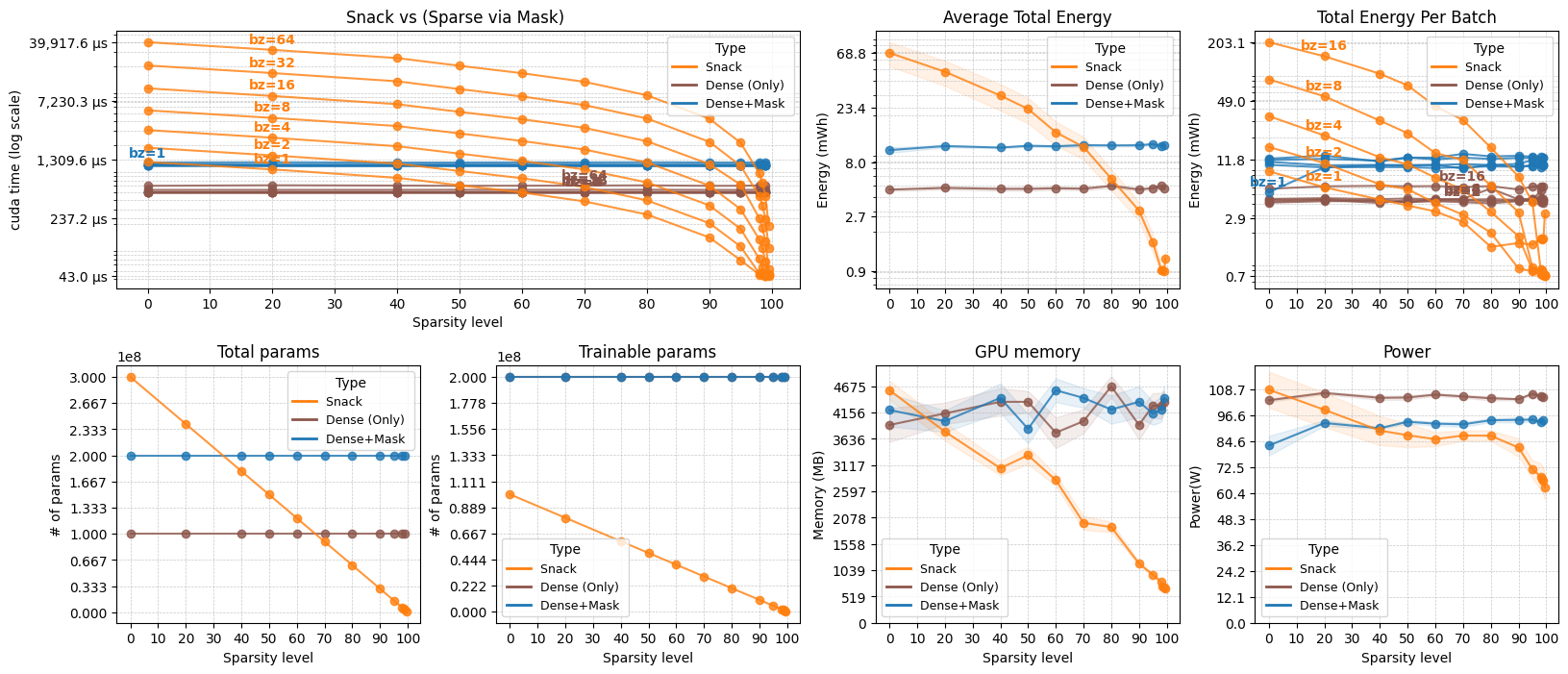}
    \caption{\textbf{Layer-level GPU profile, $10000{\times}10000$ projection ($100$ SGD iterations).}  \emph{Top row:} per-iteration CUDA time (log $\mu$s; the multiple lines per method are mini-batches $b\!\in\!\{1,2,4,8,16,32,64\}$, annotated \texttt{bz=$b$}); average total energy (mWh); total energy per batch (mWh, same $b$-sweep). \emph{Bottom row:} total parameters (sparse values + COO indices for SNACK), trainable parameters, peak GPU memory (MB), and steady-state power (W); these four panels report a single curve per method. Only SNACK's parameter, memory, and energy curves fall linearly with $1-\eta$; \texttt{Dense+Mask} stays flat because it still allocates and multiplies through a dense $W$. CUDA-time and energy axes are log-scaled because of the wide dynamic range across $\eta$.}
    \label{fig:mask+dense-vs-snack-100-gpu-profile}
\end{figure}

We replace one $10000\times 10000$ linear projection with SNACK and run $100$ SGD iterations against \emph{Dense+Mask} (the simulated-sparsity baseline) and \emph{Dense (Only)}. Figure~\ref{fig:mask+dense-vs-snack-100-gpu-profile} sweeps sparsity (and, for the two latency / per-batch energy panels, mini-batch size) across seven GPU-level metrics; Table~\ref{tab:performance_comparison_snack} pins the same numbers at batch~$4$. At $\eta=0.9$, SNACK uses only $30\%$ of Dense's parameters ($10\%$ trainable plus their COO indices) and $28\%$ of its memory; at $\eta=0.99$ it is $17\times$ faster and ${\approx}12\times$ more energy-efficient than Dense+Mask. Crucially, SNACK's parameter, memory, and energy costs scale \emph{linearly} with $1-\eta$, while Dense+Mask remains essentially flat: direct evidence that masking forfeits the savings sparsity should provide.

\begin{table*}[ht]
\footnotesize
\centering
\caption{Performance and environmental impact comparison of different implementations on a $10000{\times}10000$ matrix at batch size 4. Energy (mWh) and CO$_2$ (mg) use local emission factors. \texttt{Dense (Only)} is sparsity-independent and is reported once at the bottom as a reference (mean over the three sparsity-block runs; per-run variation is $<2\%$ on every column except memory).}
\label{tab:performance_comparison_snack}
\begin{tabular}{lcrrrrrr}
\toprule
\multirow{2}{*}{\textbf{Type}} & Sparsity & \textbf{Params} & \textbf{Memory} & \textbf{CUDA $\tau$} & \textbf{Power} & \textbf{Energy} & \textbf{CO$_2$} \\
 & \%& (k) & (MB) & ($\mu s$) & (W) & (mWh) & (mg) \\
\midrule
Dense+Mask    & 90 & 200,000 & 4371 & 1085.6 & 94.6 & 3.08 & 1.38 \\
SNACK (ours)  & 90 & \textbf{30,000} & \textbf{1173} & \textbf{484.4} & \textbf{81.7} & \textbf{0.85} & \textbf{0.38} \\
\midrule
Dense+Mask    & 95 & 200,000 & 4140 & 1085.6 & 94.7 & 3.15 & 1.42 \\
SNACK (ours)  & 95 & \textbf{15,000} & \textbf{947} & \textbf{254.6} & \textbf{71.6} & \textbf{0.45} & \textbf{0.20} \\
\midrule
Dense+Mask    & 99 & 200,000 & 4447 & 1085.3 & 94.4 & 3.07 & 1.38 \\
SNACK (ours)  & 99 & \textbf{3,000}  & \textbf{693}  & \textbf{63.6}  & \textbf{66.3} & \textbf{0.26} & \textbf{0.12} \\
\midrule
Dense (Only)  & -- & 100,000 & 4192 & 497.3  & 105.4 & 1.30 & 0.59 \\
\bottomrule
\end{tabular}
\end{table*}

\vspace{-2em}
\subsection{Numerical Equivalence and DST Accuracy}\label{subsec:exp_accuracy}
\vspace{-0.5em}

SNACK is a drop-in replacement: forward outputs and gradients match Dense+Mask up to relative error $\epsilon\!\approx\!10^{-4}$ (attributable to floating-point reordering), so any SNN trained with Dense+Mask can be swapped to SNACK without changes to the optimizer or the loss. Plugged into the DST pipeline of~\cite{lasby2023dynamic}, RigL(SNACK) matches RigL on ResNet-50/ImageNet at $\eta\in\{0.8, 0.9, 0.95, 0.99\}$ within $0.2\%$ top-1 accuracy (see Appendix~\ref{app:rigl}).

\vspace{-1em}
\subsection{Memory Headroom at Larger Layers}\label{subsec:exp_memory}
\vspace{-0.5em}

\begin{figure}[t]
    \centering
    \includegraphics[width=0.7\linewidth]{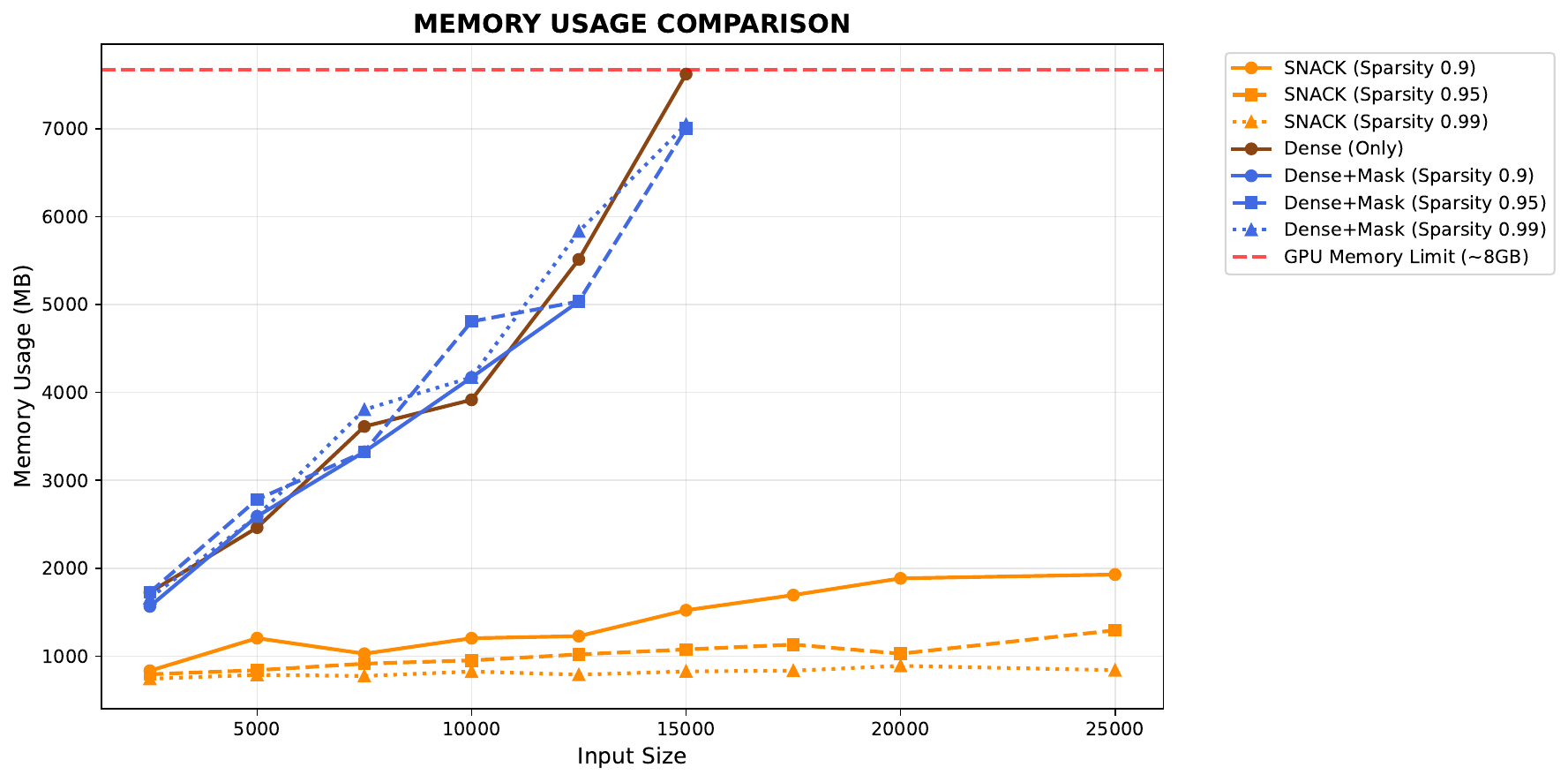}
    \caption{\textbf{Memory vs.\ input size at hidden dimension $10000$.} Dense and Dense+Mask OOM at input~$\geq 15000$; Dense+Mask at $\eta=0.9$ fails earlier ($\geq 12500$) due to lazy allocation. SNACK scales gracefully because no step ever materializes a dense $W$.}
    \label{fig:memory-usage-lines}
\end{figure}

We sweep input size at hidden dimension $10000$, recording peak memory at $\eta\in\{0.9, 0.95, 0.99\}$ and the first input size that triggers CUDA OOM (Out Of Memory) (Figure~\ref{fig:memory-usage-lines}). Dense and Dense+Mask OOM at input~$\geq 15000$ (inside the hidden-size range of modern LLMs~\cite{touvron2023llama}) because the binary mask still requires a dense $W$. SNACK avoids OOM at every sparsity tested by storing only non-zeros, fitting the same workload in roughly an order of magnitude less memory at $\eta=0.99$. Granular per-input-size profiles ($2500$\,--$12500$) and a numerical breakdown at $12500\times 10000$ are in Appendix~\ref{app:perf_bench}.

\vspace{-1em}
\section{End-to-end Experiments}
\label{sec:end_to_end_experiments}
\vspace{-0.5em}

We test SNACK end-to-end on (i) GPT-2~\cite{radford2019gpt2} dynamic sparse training and inference on WikiText-103~\cite{merity2016pointer} via HuggingFace \texttt{transformers}~\cite{wolf2020transformers}, and (ii) batch-$1$ GAMLP-style~\cite{zhang2022graph} inference on \texttt{ogbn-products}~\cite{hu2020open}. Backends: \texttt{dense} (cuBLAS), \texttt{Dense+Mask}, \texttt{SNACK(Sputnik)}; GAMLP also tests \texttt{SNACK-COO}. Sparsities are \emph{layer-local} (only \texttt{c\_fc}/\texttt{c\_proj} per GPT-2 block and the $5000{\times}5000$ GAMLP feature head are replaced; attention, embeddings, norms, LM (Language Model) head, and graph propagation stay dense), so for GPT-2 (small) $\eta{=}0.9$ removes only ${\approx}41\%$ of total parameters --- an architectural ceiling, not a kernel limit. \emph{All GPT-2 runs use a consumer-class RTX~3070 (6\,GB)} to demonstrate viability on commodity hardware where the weight-tensor memory bill is the binding constraint, so we lead on memory (and, for inference, latency). The two backends \emph{cannot be seed-aligned} --- \texttt{SNACK(Sputnik)} draws only $(1{-}\eta)mk$ non-zeros via Algorithm~\ref{alg:uni_init}, \texttt{Dense+Mask} draws Gaussian-then-mask, distinct RNG streams with no alignment hook --- so we report distributional equivalence over independent seeds, not bit-identity.

\begin{table*}[t]
    \centering
    \footnotesize
    \caption{\textbf{GPT-2 (small) dynamic sparse training across target sparsity levels.} We compare \texttt{Dense+Mask} and \texttt{SNACK(Sputnik)} (ours), reporting \emph{validation} perplexity (WikiText-103 validation split), peak GPU memory (MB), per-step energy (mJ), and per-token training latency (ms); \texttt{dense} is the no-sparsity reference. Best per-metric per-sparsity value is in \textbf{bold}.}
    \label{tab:gpt2_training_compact}
    \begin{tabular}{llrrrr}
    \toprule
    Sparsity & Method & Perplexity & Memory & Energy\_spent & Latency\_token\_ms \\
    \midrule
    0.25 & Dense+Mask & 29.92 & \textbf{2260.84} & \textbf{6916.40} & \textbf{41.64} \\
    0.25 & SNACK(Sputnik) (ours) & \textbf{29.73} & 2506.92 & 345236.72 & 582.63 \\
    \midrule
    0.50 & Dense+Mask & \textbf{36.50} & 2260.84 & \textbf{7021.94} & \textbf{41.59} \\
    0.50 & SNACK(Sputnik) (ours) & 36.89 & \textbf{2126.79} & 190115.90 & 453.23 \\
    \midrule
    0.75 & Dense+Mask & 43.08 & 2260.84 & \textbf{7127.47} & \textbf{41.53} \\
    0.75 & SNACK(Sputnik) (ours) & \textbf{43.05} & \textbf{1746.65} & 34995.07 & 332.92 \\
    \midrule
    0.85 & Dense+Mask & \textbf{62.64} & 2260.84 & \textbf{7078.19} & \textbf{41.69} \\
    0.85 & SNACK(Sputnik) (ours) & 62.72 & \textbf{1601.57} & 18627.35 & 176.56 \\
    \midrule
    0.90 & Dense+Mask & 77.63 & 2260.84 & \textbf{7054.77} & \textbf{41.72} \\
    0.90 & SNACK(Sputnik) (ours) & \textbf{77.25} & \textbf{1498.08} & 14217.54 & 127.16 \\
    \midrule
    0.95 & Dense+Mask & 103.97 & 2260.84 & \textbf{7011.71} & \textbf{41.71} \\
    0.95 & SNACK(Sputnik) (ours) & \textbf{103.16} & \textbf{1420.60} & 10171.30 & 86.13 \\
    \midrule
    0.99 & Dense+Mask & 131.90 & 2260.84 & \textbf{6978.18} & \textbf{41.76} \\
    0.99 & SNACK(Sputnik) (ours) & \textbf{130.28} & \textbf{1366.27} & 8166.11 & 70.67 \\
    \midrule
    - & dense & 29.73 & 2031.74 & 6683.58 & 39.93 \\
    \bottomrule
    \end{tabular}
    \end{table*}

\vspace{-1em}
\paragraph{GPT-2 training (Table~\ref{tab:gpt2_training_compact}): an explicit time/energy~vs.~memory trade.} \texttt{SNACK(Sputnik)} matches (and slightly better) \texttt{Dense+Mask} perplexity at every $\eta$ ($\Delta\mathrm{ppl}\!\in\![-1.62,+0.39]$ (ppl here is perplexity), due to unseedable initializer divergence) and \emph{always} cuts peak memory ($1366$ vs.\ $2261$\,MB at $\eta{=}0.99$, ${\approx}40\%$ monotone in $\eta$); the cost is per-step latency and energy, higher at low $\eta$ ($582$\,ms / $345$k\,mJ) and at parity by high $\eta$ ($71$\,ms / $8$k\,mJ between $\eta{=}0.25$ and $0.99$) as the fixed launch / format-conversion overhead amortizes against $(1{-}\eta)mnb$ arithmetic --- the trade flips toward SNACK exactly where DST operates (per-step energy ratio ${\approx}50\times{\!\rightarrow\!}{\approx}1.17\times$ across the sweep). \textbf{The point of the trade is not faster steps, it is feasibility:} the same recipe places a $124$\,M-parameter GPT-2 onto cards that cannot host \texttt{Dense+Mask} at all --- on the RTX~3070 (6\,GB) used here \texttt{SNACK(Sputnik)} at $\eta{=}0.99$ frees ${\approx}900$\,MB beyond \texttt{Dense+Mask}'s ${\approx}3.7$\,GB headroom, and the same recipe in principle drops onto a $2$\,GB-class accelerator on which \texttt{Dense+Mask} OOMs immediately.
    
    \begin{table*}[hbt]
    \centering
    \footnotesize
    \caption{\textbf{GPT-2 (small) inference across target sparsity levels at batch size~1} on a consumer-class RTX~3070 (6\,GB). We compare \texttt{Dense+Mask} and \texttt{SNACK(Sputnik)} (ours), reporting peak GPU memory (MB), per-token energy (mJ), and per-token latency (ms); \texttt{dense} is the no-sparsity reference. Best per-metric per-sparsity value is in \textbf{bold}.}
    \label{tab:gpt2_inference_compact}
    \begin{tabular}{llrrr}
    \toprule
    Sparsity & Method & Memory & Energy\_spent & Latency\_token\_ms \\
    \midrule
    0.25 & Dense+Mask              & \textbf{1255.59} & \textbf{761.16}  & \textbf{5.42} \\
    0.25 & SNACK(Sputnik) (ours)   & 1609.92          & 2237.78          & 11.91         \\
    \midrule
    0.5  & Dense+Mask              & \textbf{1255.59} & \textbf{749.48}  & \textbf{5.33} \\
    0.5  & SNACK(Sputnik) (ours)   & 1284.49          & 1507.93          & 9.31          \\
    \midrule
    0.75 & Dense+Mask              & 1255.59          & \textbf{737.79}  & \textbf{5.25} \\
    0.75 & SNACK(Sputnik) (ours)   & \textbf{959.05}  & 778.07           & 6.70          \\
    \midrule
    0.85 & Dense+Mask              & 1255.59          & 740.36           & \textbf{5.44} \\
    0.85 & SNACK(Sputnik) (ours)   & \textbf{807.59}  & \textbf{685.85}  & 6.89          \\
    \midrule
    0.9  & Dense+Mask              & 1250.87          & 730.59           & \textbf{5.33} \\
    0.9  & SNACK(Sputnik) (ours)   & \textbf{707.97}  & \textbf{627.04}  & 6.95          \\
    \midrule
    0.95 & Dense+Mask              & 1255.59          & 725.91           & \textbf{5.33} \\
    0.95 & SNACK(Sputnik) (ours)   & \textbf{632.92}  & \textbf{593.18}  & 6.82          \\
    \midrule
    0.99 & Dense+Mask              & 1255.59          & 726.93           & \textbf{5.41} \\
    0.99 & SNACK(Sputnik) (ours)   & \textbf{578.59}  & \textbf{565.64}  & 6.87          \\
    \midrule
    -    & dense                   & 1017.14          & 605.89           & \textbf{5.09} \\
    \bottomrule
    \end{tabular}
    \end{table*}

\vspace{-1em}

\paragraph{GPT-2 inference (Table~\ref{tab:gpt2_inference_compact}):}  \texttt{SNACK(Sputnik)} drops peak memory monotonically from $1610$ to $579$\,MB ($2.2\times$ over \texttt{Dense+Mask}'s flat ${\approx}1255$\,MB), keeps latency within ${\approx}1.3\times$ of dense at $\eta\geq 0.75$, and \emph{crosses into a per-token energy win} at $\eta\geq 0.85$ (${-22}\%$ at $\eta{=}0.99$): the time/energy/memory trade-off no longer trades --- SNACK is at parity (latency) or ahead (memory; energy at high $\eta$) on every axis. \textbf{Deployment consequence: the same model fits in $579$\,MB at $\eta{=}0.99$ --- under half of \texttt{Dense+Mask}'s $1255$\,MB --- freeing room for two co-resident copies, doubled context, or sub-$1$\,GB accelerators where \texttt{Dense+Mask} cannot fit.}

\begin{table*}[t]
    \centering
    \footnotesize
    \caption{\textbf{Graph-style sparse inference on a $5000{\times}5000$ GAMLP-shaped workload at batch size~1}, averaged over 2{,}000 samples after warm-up. We compare \texttt{Dense+Mask} and \texttt{SNACK(SNACK-COO)} (ours), reporting peak memory (MB), per-sample energy (mJ), and inference latency (ms); \texttt{dense} is the no-sparsity reference. Best per-metric per-sparsity value is in \textbf{bold}.}
    \label{tab:spmm_inference_compact}
    \begin{tabular}{llrrr}
    \toprule
    Method & Sparsity & Memory & Energy\_spent & Inference\_latency\_ms \\
    \midrule
    Dense+Mask & 0.9 & 408.56 & 52.79 & 1.02 \\
    SNACK(SNACK-COO) (ours) & 0.9 & \textbf{128.60} & \textbf{11.82} & \textbf{0.25} \\
    Dense+Mask & 0.95 & 408.56 & 53.63 & 1.01 \\
    SNACK(SNACK-COO) (ours) & 0.95 & \textbf{118.60} & \textbf{10.14} & \textbf{0.21} \\
    Dense+Mask & 0.99 & 309.89 & 50.86 & 1.01 \\
    SNACK(SNACK-COO) (ours) & 0.99 & \textbf{110.60} & \textbf{9.82} & \textbf{0.21} \\
    \midrule
    dense & baseline & 367.71 & 47.51 & 0.91 \\
    \bottomrule
    \end{tabular}
    \end{table*}
    
    \vspace{-1em}
\paragraph{GAMLP graph inference (Table~\ref{tab:spmm_inference_compact}).} Training runs at batch~$4096$ via SNACK(Sputnik); the trained sparse weights are then served at batch~$1$ through SNACK-COO (benchmarks are in Appendix~\ref{app:gamlp_setup}). SNACK-COO cuts peak memory $2.8\times$ and latency $4.8\times$ over \texttt{Dense+Mask} at $\eta{=}0.99$ ($110.6$ vs.\ $309.9$\,MB; $0.21$ vs.\ $1.01$\,ms), and is $3.6$--$4.3\times$ faster and $2.8$--$3.3\times$ smaller than the dense reference at every $\eta$ tested --- confirming the per-regime story: SNACK-COO wins at small leading dimension and high $\eta$, \texttt{SNACK(Sputnik)} covers larger-batch training.

\vspace{-1em}
\section{Conclusion, Limitations, and Future Work}
\vspace{-0.5em}

We presented \textbf{SNACK}, a backend-agnostic, truly sparse PyTorch layer for commodity GPUs, and \textbf{SNACK-COO}, a COO-format SpMM kernel whose batch-to-Streaming-Multiprocessor mapping targets the small-batch, high-sparsity regime current libraries underserve. The layer preserves true sparsity through forward, the sparse outer-product backward DST topology updates via the memory-cheap uniform initializer, and admits cuSPARSE / Sputnik / FlashSparse as drop-in backends. Layer-wise, this yields up to $7\times$ kernel-level speedup over Dense+Mask, up to $40\%$ peak-memory reduction in GPT-2 dynamic sparse training, and $4.8\times$ lower graph-inference latency at $99\%$ sparsity, all measured on a single \emph{consumer-class RTX~3070 (6\,GB)} chosen specifically to demonstrate viability on commodity hardware, at validation-perplexity / accuracy parity with Dense+Mask on the workloads we test. By decoupling the sparse-PyTorch surface from any single SpMM, SNACK turns each new kernel into a transparent layer-level speedup, redirecting research effort toward the kernels themselves.

\vspace{-1em}
\paragraph{Limitations.} SNACK-COO is tuned for $n\!\ll\!m$ and $\eta\!\geqslant\!0.9$ in \texttt{fp32}; outside this regime, kernels for larger batches or moderate sparsity (Sputnik, FlashSparse, BLaST) are faster, and the layer relies on backend swapping rather than a single dominant kernel. End-to-end gains are bounded architecturally: only sparsified projections recover compute (e.g.\ ${\approx}41\%$ of GPT-2-small at $\eta=0.9$).  Although SNACK, as a framework, due to its flexibility and modularity, can easily integrate Sputnik, FlashSparse, or CuSparse, its advantage on large models is investigated just for one architecture, leaving the other models largely under-explored. Yet, the potential of using it in conjunction with a large batch size is a promising research direction.

\vspace{-1em}
\paragraph{Future Work.} Immediate extensions: Tensor-Core SNACK-COO for \texttt{fp16}/\texttt{bf16}, multi-GPU SNACK with sparsity-aware collectives, and a runtime backend dispatcher. The substrate also generalises to sparse attention, MoE routers, quantization-$\times$-sparsity, and a Triton/\texttt{torch.compile} port for AMD/Intel.

\bibliographystyle{plainnat}
\bibliography{aaai2026}

\begin{thebibliography}{60}
\providecommand{\natexlab}[1]{#1}
\providecommand{\url}[1]{\texttt{#1}}
\expandafter\ifx\csname urlstyle\endcsname\relax
  \providecommand{\doi}[1]{doi: #1}\else
  \providecommand{\doi}{doi: \begingroup \urlstyle{rm}\Url}\fi

\bibitem[Agarwalla et~al.(2024)Agarwalla, Gupta, Marques, Pandit, Goin, Kurtic,
  Leong, Nguyen, Salem, Alistarh, et~al.]{agarwalla2024enabling}
Abhinav Agarwalla, Abhay Gupta, Alexandre Marques, Shubhra Pandit, Michael
  Goin, Eldar Kurtic, Kevin Leong, Tuan Nguyen, Mahmoud Salem, Dan Alistarh,
  et~al.
\newblock Enabling high-sparsity foundational {L}lama models with efficient
  pretraining and deployment.
\newblock \emph{arXiv preprint arXiv:2405.03594}, 2024.

\bibitem[Atashgahi et~al.(2022)Atashgahi, Sokar, van~der Lee, Mocanu, Mocanu,
  Veldhuis, and Pechenizkiy]{Atashgahi2022}
Zahra Atashgahi, Ghada Sokar, Tim van~der Lee, Elena Mocanu, Decebal~Constantin
  Mocanu, Raymond Veldhuis, and Mykola Pechenizkiy.
\newblock Quick and robust feature selection: the strength of energy-efficient
  sparse training for autoencoders.
\newblock \emph{Mach. Learn.}, 111\penalty0 (1):\penalty0 377–414, January
  2022.
\newblock ISSN 0885-6125.

\bibitem[Bell and Garland(2009)]{bell2009implementing}
Nathan Bell and Michael Garland.
\newblock Implementing sparse matrix-vector multiplication on
  throughput-oriented processors.
\newblock In \emph{Proceedings of the conference on high performance computing
  networking, storage and analysis}, pages 1--11, 2009.

\bibitem[Bellec et~al.(2018)Bellec, Kappel, Maass, and
  Legenstein]{bellec2018deep}
Guillaume Bellec, David Kappel, Wolfgang Maass, and Robert Legenstein.
\newblock Deep rewiring: Training very sparse deep networks.
\newblock In \emph{International Conference on Learning Representations}, 2018.

\bibitem[Bradbury et~al.(2018)Bradbury, Frostig, Hawkins, Johnson, Leary,
  Maclaurin, Necula, Paszke, Vander{P}las, Wanderman-{M}ilne, and
  Zhang]{jax2018github}
James Bradbury, Roy Frostig, Peter Hawkins, Matthew~James Johnson, Chris Leary,
  Dougal Maclaurin, George Necula, Adam Paszke, Jake Vander{P}las, Skye
  Wanderman-{M}ilne, and Qiao Zhang.
\newblock {JAX}: composable transformations of {P}ython+{N}um{P}y programs,
  2018.
\newblock URL \url{http://github.com/jax-ml/jax}.

\bibitem[Curci et~al.(2021)Curci, Mocanu, and Pechenizkiyi]{curci2021truly}
Selima Curci, Decebal~Constantin Mocanu, and Mykola Pechenizkiyi.
\newblock Truly sparse neural networks at scale.
\newblock \emph{arXiv preprint arXiv:2102.01732}, 2021.

\bibitem[Dettmers and Zettlemoyer(2019)]{dettmers2019sparse}
Tim Dettmers and Luke Zettlemoyer.
\newblock Sparse networks from scratch: Faster training without losing
  performance.
\newblock \emph{arXiv preprint arXiv:1907.04840}, 2019.

\bibitem[Evci et~al.(2020)Evci, Gale, Menick, Castro, and
  Elsen]{evci2020rigging}
Utku Evci, Trevor Gale, Jacob Menick, Pablo~Samuel Castro, and Erich Elsen.
\newblock Rigging the lottery: Making all tickets winners.
\newblock In \emph{International conference on machine learning}, pages
  2943--2952. PMLR, 2020.

\bibitem[Frankle and Carbin(2019)]{frankle2018lottery}
Jonathan Frankle and Michael Carbin.
\newblock The lottery ticket hypothesis: Finding sparse, trainable neural
  networks.
\newblock In \emph{International Conference on Learning Representations}, 2019.
\newblock URL \url{https://openreview.net/forum?id=rJl-b3RcF7}.

\bibitem[Frantar and Alistarh(2023)]{frantar2023sparsegpt}
Elias Frantar and Dan Alistarh.
\newblock Sparsegpt: Massive language models can be accurately pruned in
  one-shot.
\newblock In \emph{International conference on machine learning}, pages
  10323--10337. PMLR, 2023.

\bibitem[Frantar et~al.(2024)Frantar, Riquelme~Ruiz, Houlsby, Alistarh, and
  Evci]{FrantarICLR2024}
Elias Frantar, Carlos Riquelme~Ruiz, Neil Houlsby, Dan Alistarh, and Utku Evci.
\newblock Scaling laws for sparsely-connected foundation models.
\newblock In B.~Kim, Y.~Yue, S.~Chaudhuri, K.~Fragkiadaki, M.~Khan, and Y.~Sun,
  editors, \emph{International Conference on Representation Learning}, volume
  2024, pages 42175--42194, 2024.
\newblock URL
  \url{https://proceedings.iclr.cc/paper_files/paper/2024/file/b83abe0c353d5dddf939e3533303220f-Paper-Conference.pdf}.

\bibitem[Frostig et~al.(2018)Frostig, Johnson, and Leary]{frostig2018compiling}
Roy Frostig, Matthew~James Johnson, and Chris Leary.
\newblock Compiling machine learning programs via high-level tracing.
\newblock \emph{Systems for Machine Learning}, 4\penalty0 (9):\penalty0 1--3,
  2018.

\bibitem[Gale et~al.(2019)Gale, Elsen, and Hooker]{gale2019state}
Trevor Gale, Erich Elsen, and Sara Hooker.
\newblock The state of sparsity in deep neural networks.
\newblock \emph{arXiv preprint arXiv:1902.09574}, 2019.

\bibitem[Gale et~al.(2020)Gale, Zaharia, Young, and Elsen]{gale2020sparse}
Trevor Gale, Matei Zaharia, Cliff Young, and Erich Elsen.
\newblock Sparse {GPU} kernels for deep learning.
\newblock In \emph{SC20: International Conference for High Performance
  Computing, Networking, Storage and Analysis}, pages 1--14. IEEE, 2020.

\bibitem[Gilbert et~al.(1992)Gilbert, Moler, and Schreiber]{gilbert1992sparse}
John~R Gilbert, Cleve Moler, and Robert Schreiber.
\newblock Sparse matrices in matlab: Design and implementation.
\newblock \emph{SIAM journal on matrix analysis and applications}, 13\penalty0
  (1):\penalty0 333--356, 1992.

\bibitem[Graesser et~al.(2022)Graesser, Evci, Elsen, and
  Castro]{graesser2022state}
Laura Graesser, Utku Evci, Erich Elsen, and Pablo~Samuel Castro.
\newblock The state of sparse training in deep reinforcement learning.
\newblock In \emph{International Conference on Machine Learning}, pages
  7766--7792. PMLR, 2022.

\bibitem[Hagan et~al.(1997)Hagan, Demuth, and Beale]{hagan1997neural}
Martin~T Hagan, Howard~B Demuth, and Mark Beale.
\newblock \emph{Neural network design}.
\newblock PWS Publishing Co., 1997.

\bibitem[Hassibi et~al.(1993)Hassibi, Stork, and Wolff]{hassibi1993optimal}
Babak Hassibi, David~G Stork, and Gregory~J Wolff.
\newblock Optimal brain surgeon and general network pruning.
\newblock In \emph{IEEE international conference on neural networks}, pages
  293--299. IEEE, 1993.

\bibitem[Hoefler et~al.(2021)Hoefler, Alistarh, Ben-Nun, Dryden, and
  Peste]{hoefler2021jmlr}
Torsten Hoefler, Dan Alistarh, Tal Ben-Nun, Nikoli Dryden, and Alexandra Peste.
\newblock Sparsity in deep learning: pruning and growth for efficient inference
  and training in neural networks.
\newblock \emph{J. Mach. Learn. Res.}, 22\penalty0 (1), January 2021.
\newblock ISSN 1532-4435.

\bibitem[Hu et~al.(2020)Hu, Fey, Zitnik, Dong, Ren, Liu, Catasta, and
  Leskovec]{hu2020open}
Weihua Hu, Matthias Fey, Marinka Zitnik, Yuxiao Dong, Hongyu Ren, Bowen Liu,
  Michele Catasta, and Jure Leskovec.
\newblock Open graph benchmark: Datasets for machine learning on graphs.
\newblock \emph{Advances in neural information processing systems},
  33:\penalty0 22118--22133, 2020.

\bibitem[Im(2000)]{im2000optimizing}
Eun-Jin Im.
\newblock \emph{Optimizing the performance of sparse matrix-vector
  multiplication}.
\newblock University of California, Berkeley, 2000.

\bibitem[Lasby et~al.(2023)Lasby, Golubeva, Evci, Nica, and
  Ioannou]{lasby2023dynamic}
Mike Lasby, Anna Golubeva, Utku Evci, Mihai Nica, and Yani Ioannou.
\newblock Dynamic sparse training with structured sparsity.
\newblock \emph{arXiv preprint arXiv:2305.02299}, 2023.

\bibitem[Lasby et~al.(2024)Lasby, Golubeva, Evci, Nica, and
  Ioannou]{lasby2024dynamicsparsetrainingstructured}
Mike Lasby, Anna Golubeva, Utku Evci, Mihai Nica, and Yani Ioannou.
\newblock Dynamic sparse training with structured sparsity, 2024.
\newblock URL \url{https://arxiv.org/abs/2305.02299}.

\bibitem[LeCun et~al.(1989)LeCun, Denker, and Solla]{NIPS1989_6c9882bb}
Yann LeCun, John Denker, and Sara Solla.
\newblock Optimal brain damage.
\newblock In D.~Touretzky, editor, \emph{Advances in Neural Information
  Processing Systems}, volume~2. Morgan-Kaufmann, 1989.
\newblock URL
  \url{https://proceedings.neurips.cc/paper_files/paper/1989/file/6c9882bbac1c7093bd25041881277658-Paper.pdf}.

\bibitem[Lee et~al.(2019)Lee, Ajanthan, and Torr]{lee2019snip}
Namhoon Lee, Thalaiyasingam Ajanthan, and Philip Torr.
\newblock {SNIP}: Single-shot network pruning based on connection sensitivity.
\newblock In \emph{International Conference on Learning Representations}, 2019.
\newblock URL \url{https://openreview.net/forum?id=B1VZqjAcYX}.

\bibitem[Li et~al.(2023)Li, Orr, Ohan, Murray, Sanders, Beker, Masters,
  et~al.]{li2023popsparse}
Zhiyi Li, Douglas Orr, Valeriu Ohan, Tom Murray, Adam Sanders, Deniz Beker,
  Dominic Masters, et~al.
\newblock Popsparse: Accelerated block sparse matrix multiplication on ipu.
\newblock \emph{arXiv preprint arXiv:2303.16999}, 2023.

\bibitem[Liang et~al.(2021)Liang, Glossner, Wang, Shi, and
  Zhang]{liang2021pruning}
Tailin Liang, John Glossner, Lei Wang, Shaobo Shi, and Xiaotong Zhang.
\newblock Pruning and quantization for deep neural network acceleration: A
  survey.
\newblock \emph{Neurocomputing}, 461:\penalty0 370--403, 2021.

\bibitem[Liu et~al.(2020)Liu, Xu, Shi, Cheung, and So]{liu2020dynamic}
Junjie Liu, Zhe Xu, Runbin Shi, Ray~CC Cheung, and Hayden~KH So.
\newblock Dynamic sparse training: Find efficient sparse network from scratch
  with trainable masked layers.
\newblock \emph{arXiv preprint arXiv:2005.06870}, 2020.

\bibitem[Liu et~al.(2021)Liu, Yin, Mocanu, and Pechenizkiy]{pmlr-v139-liu21y}
Shiwei Liu, Lu~Yin, Decebal~Constantin Mocanu, and Mykola Pechenizkiy.
\newblock Do we actually need dense over-parameterization? in-time
  over-parameterization in sparse training.
\newblock In Marina Meila and Tong Zhang, editors, \emph{Proceedings of the
  38th International Conference on Machine Learning}, volume 139 of
  \emph{Proceedings of Machine Learning Research}, pages 6989--7000. PMLR,
  18--24 Jul 2021.
\newblock URL \url{https://proceedings.mlr.press/v139/liu21y.html}.

\bibitem[Luo et~al.(2026)Luo, Zhang, Hu, Zhang, Wang, Su, Sun, Liang, and
  Zhang]{luo2026sparse}
Sijia Luo, Xiaokang Zhang, Yuxuan Hu, Bohan Zhang, Ke~Wang, Jinbo Su, Mengshu
  Sun, Lei Liang, and Jing Zhang.
\newblock Sparse-rl: Breaking the memory wall in llm reinforcement learning via
  stable sparse rollouts.
\newblock \emph{arXiv preprint arXiv:2601.10079}, 2026.

\bibitem[Merity et~al.(2017)Merity, Xiong, Bradbury, and
  Socher]{merity2016pointer}
Stephen Merity, Caiming Xiong, James Bradbury, and Richard Socher.
\newblock Pointer sentinel mixture models.
\newblock In \emph{International Conference on Learning Representations
  (ICLR)}, 2017.

\bibitem[Merrill and Garland(2016)]{merrill2016merge}
Duane Merrill and Michael Garland.
\newblock Merge-based parallel sparse matrix-vector multiplication.
\newblock In \emph{SC'16: Proceedings of the International Conference for High
  Performance Computing, Networking, Storage and Analysis}, pages 678--689.
  IEEE, 2016.

\bibitem[Mocanu et~al.(2018)Mocanu, Mocanu, Stone, Nguyen, Gibescu, and
  Liotta]{mocanu2018scalable}
Decebal~Constantin Mocanu, Elena Mocanu, Peter Stone, Phuong~H Nguyen,
  Madeleine Gibescu, and Antonio Liotta.
\newblock Scalable training of artificial neural networks with adaptive sparse
  connectivity inspired by network science.
\newblock \emph{Nature communications}, 9\penalty0 (1):\penalty0 2383, 2018.

\bibitem[Mocanu et~al.(2021)Mocanu, Mocanu, Pinto, Curci, Nguyen, Gibescu,
  Ernst, and Vale]{Mocanu2021amaas}
Decebal~Constantin Mocanu, Elena Mocanu, Tiago Pinto, Selima Curci, Phuong~H.
  Nguyen, Madeleine Gibescu, Damien Ernst, and Zita~A. Vale.
\newblock Sparse training theory for scalable and efficient agents.
\newblock In \emph{Proceedings of the 20th International Conference on
  Autonomous Agents and MultiAgent Systems}, AAMAS '21, page 34–38, Richland,
  SC, 2021. International Foundation for Autonomous Agents and Multiagent
  Systems.
\newblock ISBN 9781450383073.

\bibitem[Mostafa and Wang(2019)]{mostafa2019parameter}
Hesham Mostafa and Xin Wang.
\newblock Parameter efficient training of deep convolutional neural networks by
  dynamic sparse reparameterization.
\newblock In \emph{International Conference on Machine Learning}, pages
  4646--4655. PMLR, 2019.

\bibitem[Mu{\~n}oz-Mart{\'\i}nez et~al.(2023)Mu{\~n}oz-Mart{\'\i}nez, Garg,
  Pellauer, Abell{\'a}n, Acacio, and Krishna]{munoz2023flexagon}
Francisco Mu{\~n}oz-Mart{\'\i}nez, Raveesh Garg, Michael Pellauer, Jos{\'e}~L
  Abell{\'a}n, Manuel~E Acacio, and Tushar Krishna.
\newblock Flexagon: A multi-dataflow sparse-sparse matrix multiplication
  accelerator for efficient dnn processing.
\newblock In \emph{Proceedings of the 28th ACM International Conference on
  Architectural Support for Programming Languages and Operating Systems, Volume
  3}, pages 252--265, 2023.

\bibitem[Naumov et~al.(2010)Naumov, Chien, Vandermersch, and
  Kapasi]{naumov2010cusparse}
Maxim Naumov, L~Chien, Philippe Vandermersch, and Ujval Kapasi.
\newblock Cusparse library: A set of basic linear algebra subroutines for
  sparse matrices.
\newblock In \emph{GPU Technology Conference}, number~5, page~17, 2010.

\bibitem[Nishino and Loomis(2017)]{nishino2017cupy}
ROYUD Nishino and Shohei Hido~Crissman Loomis.
\newblock Cupy: A numpy-compatible library for nvidia gpu calculations.
\newblock \emph{31st confernce on neural information processing systems},
  151\penalty0 (7), 2017.

\bibitem[Okanovic et~al.(2025)Okanovic, Deshmukh, Kwasniewski, Zhu, Fujii,
  Fatima, Besta, Katayama, Honda, Nagasaka, et~al.]{okanovic2025blast}
Patrik Okanovic, Sameer Deshmukh, Grzegorz Kwasniewski, Yi~Zhu, Haruto Fujii,
  Sakina Fatima, Maciej Besta, Kentaro Katayama, Takumi Honda, Yusuke Nagasaka,
  et~al.
\newblock Blast: High performance inference and pretraining using block sparse
  transformers.
\newblock \emph{arXiv preprint arXiv:2507.03117}, 2025.

\bibitem[Paszke et~al.(2019)Paszke, Gross, Massa, Lerer, Bradbury, Chanan,
  Killeen, Lin, Gimelshein, Antiga, Desmaison, K\"{o}pf, Yang, DeVito, Raison,
  Tejani, Chilamkurthy, Steiner, Fang, Bai, and
  Chintala]{10.5555/3454287.3455008}
Adam Paszke, Sam Gross, Francisco Massa, Adam Lerer, James Bradbury, Gregory
  Chanan, Trevor Killeen, Zeming Lin, Natalia Gimelshein, Luca Antiga, Alban
  Desmaison, Andreas K\"{o}pf, Edward Yang, Zach DeVito, Martin Raison, Alykhan
  Tejani, Sasank Chilamkurthy, Benoit Steiner, Lu~Fang, Junjie Bai, and Soumith
  Chintala.
\newblock \emph{PyTorch: an imperative style, high-performance deep learning
  library}.
\newblock Curran Associates Inc., Red Hook, NY, USA, 2019.

\bibitem[Radford et~al.(2019)Radford, Wu, Child, Luan, Amodei, and
  Sutskever]{radford2019gpt2}
Alec Radford, Jeffrey Wu, Rewon Child, David Luan, Dario Amodei, and Ilya
  Sutskever.
\newblock Language models are unsupervised multitask learners.
\newblock Technical report, OpenAI, 2019.

\bibitem[Saad(2003)]{saad2003iterative}
Yousef Saad.
\newblock \emph{Iterative methods for sparse linear systems}.
\newblock SIAM, 2003.

\bibitem[Schultheis and Babbar(2023)]{schultheis2023towards}
Erik Schultheis and Rohit Babbar.
\newblock Towards memory-efficient training for extremely large output
  spaces--learning with 670k labels on a single commodity gpu.
\newblock In \emph{Joint European Conference on Machine Learning and Knowledge
  Discovery in Databases}, pages 689--704. Springer, 2023.

\bibitem[Schwartz et~al.(2020)Schwartz, Dodge, Smith, and
  Etzioni]{schwartz2020green}
Roy Schwartz, Jesse Dodge, Noah~A Smith, and Oren Etzioni.
\newblock Green ai.
\newblock \emph{Communications of the ACM}, 63\penalty0 (12):\penalty0 54--63,
  2020.

\bibitem[Shi et~al.(2025)Shi, Li, Xu, Fu, Wang, and Wu]{shi2025flashsparse}
Jinliang Shi, Shigang Li, Youxuan Xu, Rongtian Fu, Xueying Wang, and Tong Wu.
\newblock {FlashSparse}: Minimizing computation redundancy for fast sparse
  matrix multiplications on tensor cores.
\newblock In \emph{Proceedings of the 30th ACM SIGPLAN Annual Symposium on
  Principles and Practice of Parallel Programming (PPoPP)}, 2025.

\bibitem[Strubell et~al.(2019)Strubell, Ganesh, and
  McCallum]{strubell2019energy}
Emma Strubell, Ananya Ganesh, and Andrew McCallum.
\newblock Energy and policy considerations for deep learning in nlp.
\newblock In \emph{Proceedings of the 57th Annual Meeting of the Association
  for Computational Linguistics}, pages 3645--3650, 2019.

\bibitem[Tanaka et~al.(2020)Tanaka, Kunin, Yamins, and
  Ganguli]{tanaka2020pruning}
Hidenori Tanaka, Daniel Kunin, Daniel~L. Yamins, and Surya Ganguli.
\newblock Pruning neural networks without any data by iteratively conserving
  synaptic flow.
\newblock In \emph{Advances in Neural Information Processing Systems},
  volume~33, pages 6377--6389, 2020.

\bibitem[Thangarasa et~al.(2023{\natexlab{a}})Thangarasa, Gupta, Marshall, Li,
  Leong, DeCoste, Lie, and Saxena]{thangarasa2023spdf}
Vithursan Thangarasa, Abhay Gupta, William Marshall, Tianda Li, Kevin Leong,
  Dennis DeCoste, Sean Lie, and Shreyas Saxena.
\newblock Spdf: Sparse pre-training and dense fine-tuning for large language
  models.
\newblock In \emph{Uncertainty in Artificial Intelligence}, pages 2134--2146.
  PMLR, 2023{\natexlab{a}}.

\bibitem[Thangarasa et~al.(2023{\natexlab{b}})Thangarasa, Saxena, Gupta, and
  Lie]{thangarasa2023sparse}
Vithursan Thangarasa, Shreyas Saxena, Abhay Gupta, and Sean Lie.
\newblock Sparse-ift: Sparse iso-flop transformations for maximizing training
  efficiency.
\newblock \emph{arXiv preprint arXiv:2303.11525}, 2023{\natexlab{b}}.

\bibitem[Touvron et~al.(2023)Touvron, Lavril, Izacard, Martinet, Lachaux,
  Lacroix, Rozi{\`e}re, Goyal, Hambro, Azhar, et~al.]{touvron2023llama}
Hugo Touvron, Thibaut Lavril, Gautier Izacard, Xavier Martinet, Marie-Anne
  Lachaux, Timoth{\'e}e Lacroix, Baptiste Rozi{\`e}re, Naman Goyal, Eric
  Hambro, Faisal Azhar, et~al.
\newblock Llama: Open and efficient foundation language models.
\newblock \emph{arXiv preprint arXiv:2302.13971}, 2023.

\bibitem[Wang et~al.(2025)Wang, Lin, Wei, Gao, and Ji]{wang2025optimizing}
Yizhuo Wang, Hongpeng Lin, Bingxin Wei, Jianhua Gao, and Weixing Ji.
\newblock Optimizing general sparse matrix-matrix multiplication on the gpu.
\newblock \emph{ACM Transactions on Architecture and Code Optimization},
  22\penalty0 (4):\penalty0 1--25, 2025.

\bibitem[Wen et~al.(2016)Wen, Wu, Wang, Chen, and Li]{wen2016learning}
Wei Wen, Chunpeng Wu, Yandan Wang, Yiran Chen, and Hai Li.
\newblock Learning structured sparsity in deep neural networks.
\newblock \emph{Advances in neural information processing systems}, 29, 2016.

\bibitem[Wolf et~al.(2020)Wolf, Debut, Sanh, Chaumond, Delangue, Moi, Cistac,
  Rault, Louf, Funtowicz, Davison, Shleifer, von Platen, Ma, Jernite, Plu, Xu,
  Le~Scao, Gugger, Drame, Lhoest, and Rush]{wolf2020transformers}
Thomas Wolf, Lysandre Debut, Victor Sanh, Julien Chaumond, Clement Delangue,
  Anthony Moi, Pierric Cistac, Tim Rault, R{\'e}mi Louf, Morgan Funtowicz, Joe
  Davison, Sam Shleifer, Patrick von Platen, Clara Ma, Yacine Jernite, Julien
  Plu, Canwen Xu, Teven Le~Scao, Sylvain Gugger, Mariama Drame, Quentin Lhoest,
  and Alexander~M. Rush.
\newblock Transformers: State-of-the-art natural language processing.
\newblock In \emph{Proceedings of the 2020 Conference on Empirical Methods in
  Natural Language Processing: System Demonstrations}, pages 38--45, 2020.

\bibitem[Yang et~al.(2018)Yang, Bulu{\c{c}}, and Owens]{yang2018design}
Carl Yang, Ayd{\i}n Bulu{\c{c}}, and John~D Owens.
\newblock Design principles for sparse matrix multiplication on the gpu.
\newblock In \emph{European Conference on Parallel Processing}, pages 672--687.
  Springer, 2018.

\bibitem[Yuan et~al.(2021)Yuan, Ma, Niu, Li, Kong, Liu, Gong, Zhan, He, Jin,
  et~al.]{yuan2021mest}
Geng Yuan, Xiaolong Ma, Wei Niu, Zhengang Li, Zhenglun Kong, Ning Liu, Yifan
  Gong, Zheng Zhan, Chaoyang He, Qing Jin, et~al.
\newblock {MEST}: Accurate and fast memory-economic sparse training framework
  on the edge.
\newblock In \emph{Advances in Neural Information Processing Systems},
  volume~34, pages 20838--20850, 2021.

\bibitem[Zhang et~al.(2021)Zhang, Attaluri, Emer, and Sanchez]{zhang2021gamma}
Guowei Zhang, Nithya Attaluri, Joel~S Emer, and Daniel Sanchez.
\newblock Gamma: Leveraging gustavson’s algorithm to accelerate sparse matrix
  multiplication.
\newblock In \emph{Proceedings of the 26th ACM International Conference on
  Architectural Support for Programming Languages and Operating Systems}, pages
  687--701, 2021.

\bibitem[Zhang et~al.(2022)Zhang, Yin, Sheng, Li, Ouyang, Li, Tao, Yang, and
  Cui]{zhang2022graph}
Wentao Zhang, Ziqi Yin, Zeang Sheng, Yang Li, Wen Ouyang, Xiaosen Li, Yangyu
  Tao, Zhi Yang, and Bin Cui.
\newblock Graph attention multi-layer perceptron.
\newblock In \emph{Proceedings of the 28th ACM SIGKDD Conference on Knowledge
  Discovery and Data Mining}, pages 4560--4570, 2022.

\bibitem[Zhang et~al.(2020)Zhang, Wang, Han, and Dally]{9065428}
Zhekai Zhang, Hanrui Wang, Song Han, and William~J. Dally.
\newblock Sparch: Efficient architecture for sparse matrix multiplication.
\newblock In \emph{2020 IEEE International Symposium on High Performance
  Computer Architecture (HPCA)}, pages 261--274, 2020.
\newblock \doi{10.1109/HPCA47549.2020.00030}.

\bibitem[Zhao et~al.(2024)Zhao, Zhang, Chen, Wang, Anandkumar, and
  Tian]{zhao2024galore}
Jiawei Zhao, Zhenyu Zhang, Beidi Chen, Zhangyang Wang, Anima Anandkumar, and
  Yuandong Tian.
\newblock {GaLore}: Memory-efficient {LLM} training by gradient low-rank
  projection.
\newblock In \emph{Forty-first International Conference on Machine Learning},
  2024.

\bibitem[Zhu and Gupta(2017)]{zhu2017prune}
Michael Zhu and Suyog Gupta.
\newblock To prune, or not to prune: exploring the efficacy of pruning for
  model compression.
\newblock \emph{arXiv preprint arXiv:1710.01878}, 2017.

\end{thebibliography}

\clearpage

\appendix
\onecolumn

\section*{Appendix}
\section{DST Background}
Sparse training of deep neural networks, also known as dynamic sparse training (DST), encompasses methods that maintain sparsity throughout training by starting from a sparse initialization and iteratively updating the network's connectivity. During training, a fraction of weights is periodically removed and regrown, allowing the sparse structure to adapt to the learning dynamics. Recent work~\cite{gale2019state, hoefler2021jmlr, Mocanu2021amaas, graesser2022state, FrantarICLR2024} provides a comprehensive overview of DST algorithms, theoretical foundations, and practical trade-offs, demonstrating their potential to scale deep learning efficiently without significant loss in performance.

\section{Sparse Matrices \& Sparse Operations}\label{app:sparse_ops}
Sparse Matrices are the cornerstone of many DST and Pruning methods. Sparse Matrices are data structures containing compact data representations of a dense matrix with missing elements (generally represented by zeroes).

Sparse matrix multiplication (SpMM) is a popular operation in many engineering applications. It is especially crucial when having large data structures. Experiments added early on by \citep{gilbert1992sparse} in MATLAB explored having sparse vectors in each row.
Assuming we have the following matrix:
\[
A = \begin{bmatrix}
0 & 0 & 3 \\
4 & 0 & 0 \\
0 & 5 & 0 \\
\end{bmatrix}
\]
Sparse matrix storage data structure is essential in optimizing memory and computational resources. There are many structures for SpM representation. A prevalent structure is the Compressed Sparse Row (CSR) format. It is used for optimized row-wise storage, reducing memory overhead since memory is optimized to be accessed in contiguous segments \citep{saad2003iterative}. A similar format is Compressed Sparse Column (CSC), which is optimized for column-wise storage. The Coordinate List format (COO) is a simple representation of the storage data structure of {(SpM)}, where non-zero values and their coordinates are explicitly stored \citep{bell2009implementing}. Examples on how each representation is shown in Table \ref{tab:sparse-formats}.
\begin{table}[ht]
\centering
\caption{Sparse matrix representations in COO, CSR, and CSC formats.}
\label{tab:sparse-formats}
\begin{tabular}{llll}
\toprule
\textbf{Format} & \textbf{Data} & \textbf{Indices} & \textbf{Index Pointer} \\
\midrule
COO & [3, 4, 5] & row = [0, 1, 2] & col = [2, 0, 1] \\
CSR & [3, 4, 5] & col = [2, 0, 1] & indptr = [0, 1, 2, 3] \\
CSC & [4, 5, 3] & row = [1, 2, 0] & indptr = [0, 1, 2, 3] \\
\bottomrule
\end{tabular}
\end{table}

Another format is the Block Sparse Row (BSR). It extends CSR by grouping non-zero elements into blocks. Blocks are usually $n\times n$ in size, and if a block does not contain a non-zero element, it gets ignored. It is less efficient, memory-wise, than other formats but more computationally fast as it improves cache efficiency and parallelism \citep{im2000optimizing}.

There are many theoretical architectures for sparse accelerators, such as SpArch, which features a fixed outer product dataflow. Flexagon advances this by supporting multiple dataflows (inner product, outer product) within a reconfigurable architecture \cite{9065428, munoz2023flexagon}. Gamma \cite{zhang2021gamma} leverages Gustavson's algorithm with specialized fiber cache and merger hardware to accelerate sparse matrix multiplication. Recent work by \cite{schultheis2023towards} demonstrates custom CUDA kernels for uniform sparse matrix multiplication in extreme classification scenarios, where each output neuron receives exactly the same number of inputs, achieving memory-efficient training with 500K labels on commodity GPUs.

\section{Experiment within Condensed Sparsity Training Phase}\label{app:rigl}
The flexibility of the SNACK layer allows for fast integration into any dynamic sparse training pipeline. In this section, we provide results integrating the SNACK layer into a CNN-based architecture. Recently, \cite{lasby2024dynamicsparsetrainingstructured} showed that DST can be effectively applied to computer vision tasks under a structured sparsity setup, achieving increased accuracy compared to its dense counterpart. The main motivation behind choosing to integrate SNACK with this work is to show how SNACK is integrable with any DST-based work and it is worthwhile to use it for energy cost reduction and restricted memory spaces environments.

Starting from their reported results, we introduce RigL(SNACK) to train a ResNet-50 on ImageNet. In order to do so, we replaced the last linear layer of ResNet-50 used in the dynamic sparse training regime via RigL with our proposed SNACK layer for the forward pass and backpropagation. The layer replacement was directly performed in the code provided by~\cite{lasby2024dynamicsparsetrainingstructured}. We ran all the experiments five times and report the mean and standard deviation in Table~\ref{tab:rigl_resnet50}.
\begin{table}
\centering
\caption{Top-1 ImageNet test accuracy of ResNet-50 trained with RigL at high sparsities. RigL* is the method used by \cite{lasby2024dynamicsparsetrainingstructured}; RigL (SNACK) is the same method but using SNACK instead of a dense layer with masked weights.}
\label{tab:rigl_resnet50}
\begin{tabular}{cccc}
\toprule
\textbf{Sparsity } & \textbf{RigL*}  &\textbf{RigL (SNACK) (ours)} & \textbf{Dense} \\
 (\%)& & & \textbf{ ResNet-50}\\
\midrule
80 & $74.9 \pm 0.1$ & $74.7 \pm 0.2$ &  \\
90 & $72.8 \pm 0.1$ & $72.8 \pm 0.1$ & \multirow{2}{*}{$76.7 $}\\
95 & $69.6 \pm 0.1$ & $69.6 \pm 0.2$ & \multirow{2}{*}{$\pm 0.1 $}\\
99 & $51.4 \pm 0.2$ & $51.4 \pm 0.3$ & \\
\bottomrule
\end{tabular}
\end{table}
It is worth highlighting that the sparse matrix multiplication using SNACK produces identical gradients to sparse-via-mask, which are used in the training phase alongside RigL. This should be reflected in exact results in terms of accuracy between both implementations. However, as can be seen in Table~\ref{tab:rigl_resnet50}, there are slight differences, due to float operations being handled differently.
\begin{table}
\centering
\caption{Energy consumption (in kWh) of ResNet-50 trained with RigL at various sparsities. RigL* are from \cite{lasby2024dynamicsparsetrainingstructured}; RigL (SNACK) replaces dense masked layers with truly sparse CUDA kernels.}
\label{tab:rigl_energy}
\begin{tabular}{cccc}
\toprule
\textbf{Sparsity} & \textbf{RigL*} & \textbf{RigL } &  \textbf{Dense}\\
 \textbf{\%}& &\textbf{(SNACK) (ours)} & \textbf{ResNet-50} \\
\midrule
80  & $5.170 \pm 0.007$ & $5.217 \pm 0.010 $& $4.26 \pm 0.003$ \\
90  & $5.172 \pm 0.005$ & $5.045 \pm 0.010$ & $4.25 \pm 0.004$ \\
95  & $5.156 \pm 0.006$  & $5.033 \pm 0.015$  & $4.24 \pm 0.003$ \\
99   & $ 5.112 \pm 0.007$ & $5.017 \pm 0.020$ & $4.25 \pm 0.003$ \\
\bottomrule
\end{tabular}
\end{table}
Furthermore, we investigate their energy consumption (see Table~\ref{tab:rigl_energy}). We observed an improvement in energy consumption of RigL(SNACK) versus RigL(via mask) for sparsities above $90\%$ (i.e $90\%$, $95\%$, and $99\%$). Of course, the small reduction in energy consumption is well calibrated with the expected impact of sparsifying only the last layer computation over the total training requirements in ResNet-50.

\section{Performance benchmarks using various SpM sizes}\label{app:perf_bench}

\subsection{Effects of different hidden layer sizes}

\begin{figure}[H]
    \centering
    \includegraphics[width=\linewidth]{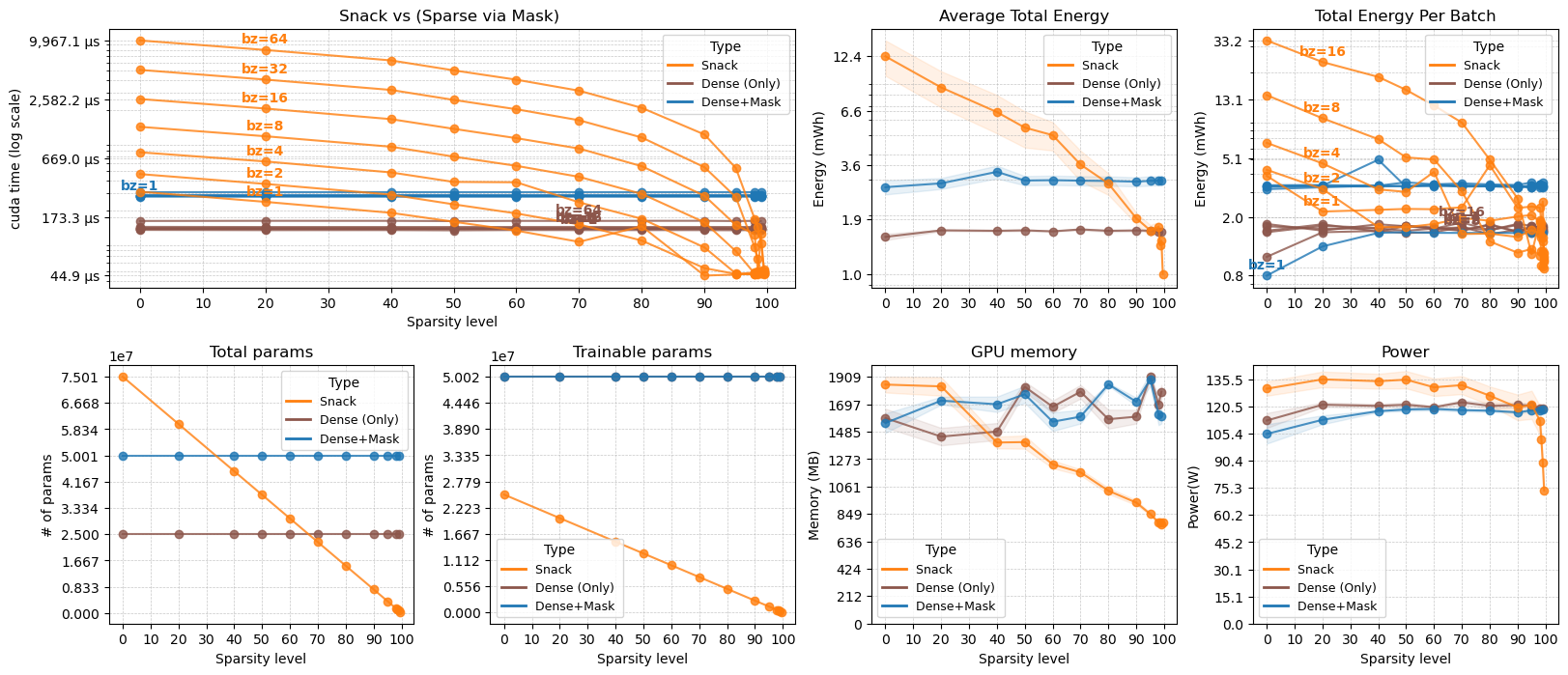}
    \caption{SNACK performance when SNACK layer dimension is 2500x10000}
    \label{fig:mask+dense-vs-snack-2500x10000-gpu-profile}
\end{figure}

\begin{figure}[H]
    \centering
    \includegraphics[width=\linewidth]{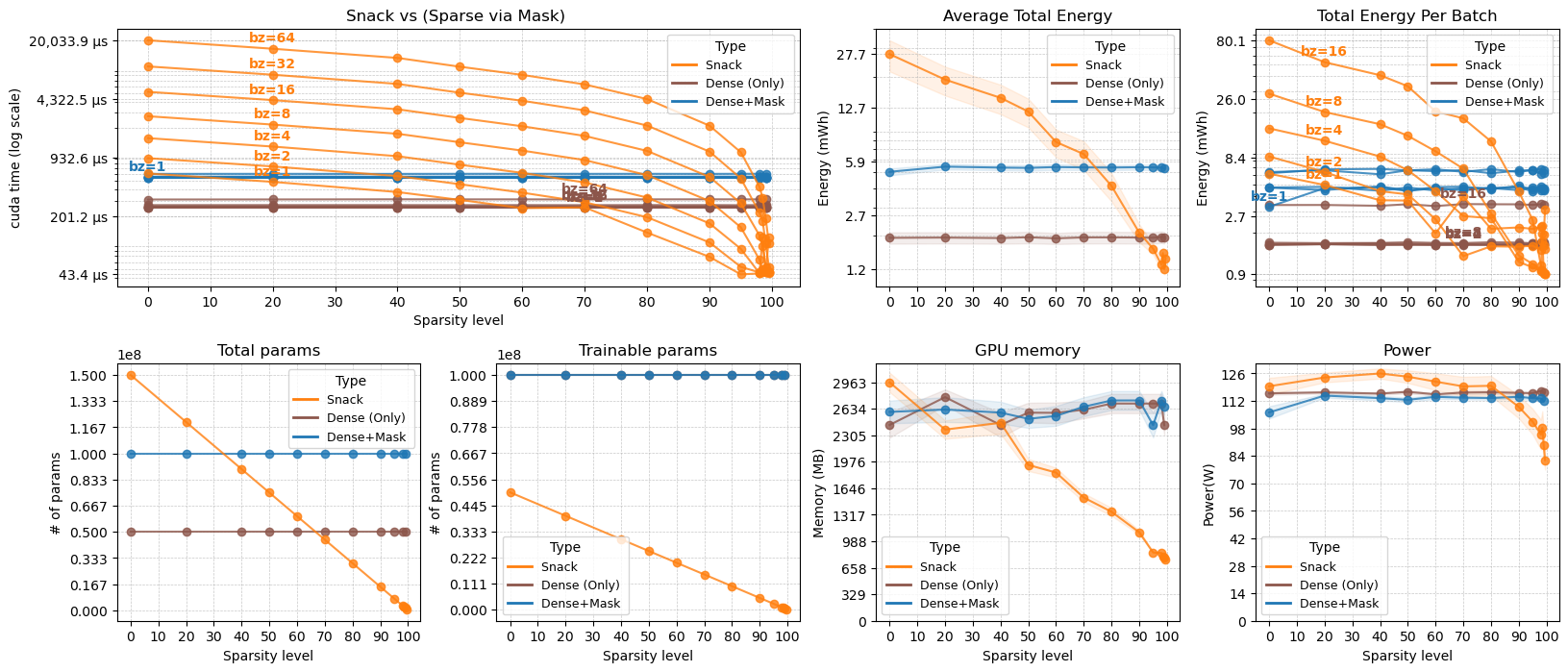}
    \caption{SNACK performance when SNACK layer dimension is 5000x10000}
    \label{fig:mask+dense-vs-snack-5000x10000-gpu-profile}
\end{figure}

\begin{figure}[H]
    \centering
    \includegraphics[width=\linewidth]{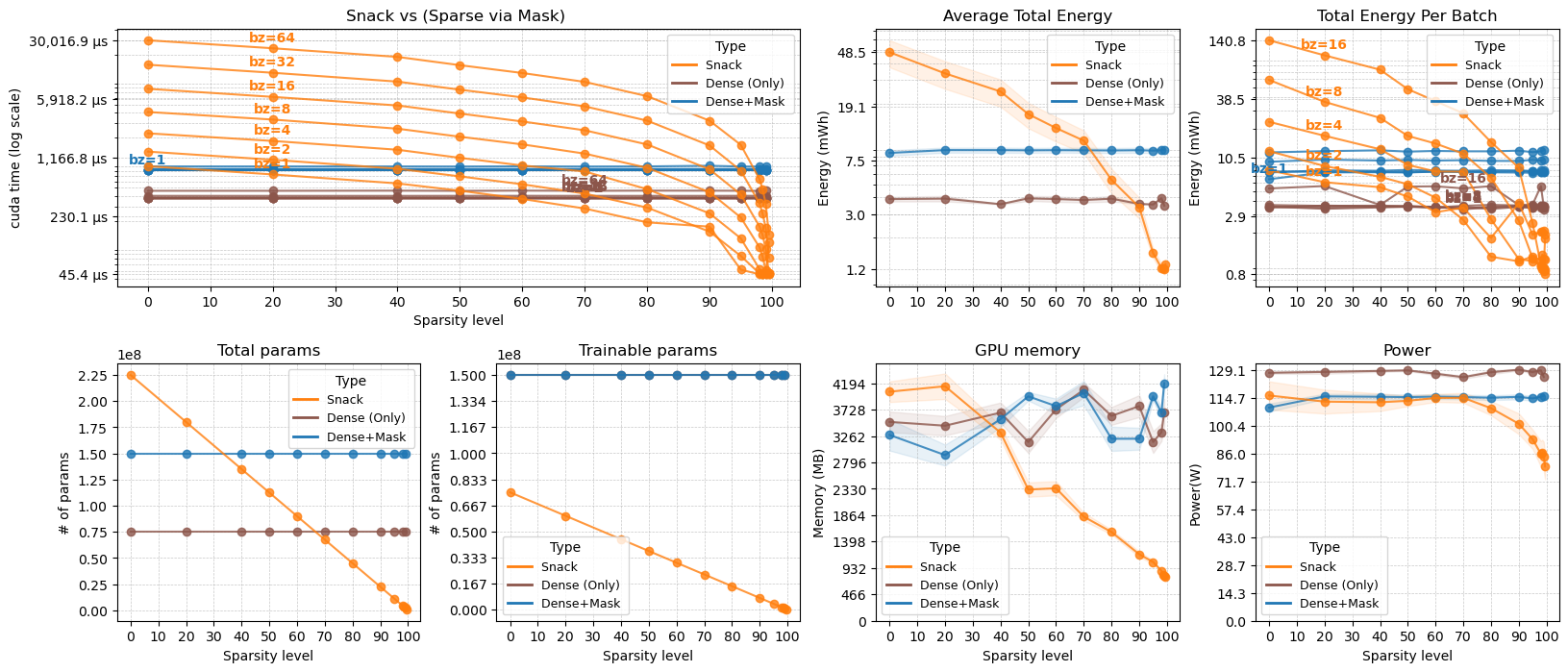}
    \caption{SNACK performance when SNACK layer dimension is 7500x10000}
    \label{fig:mask+dense-vs-snack-7500x10000-gpu-profile}
\end{figure}

\begin{figure}[H]
    \centering
    \includegraphics[width=\linewidth]{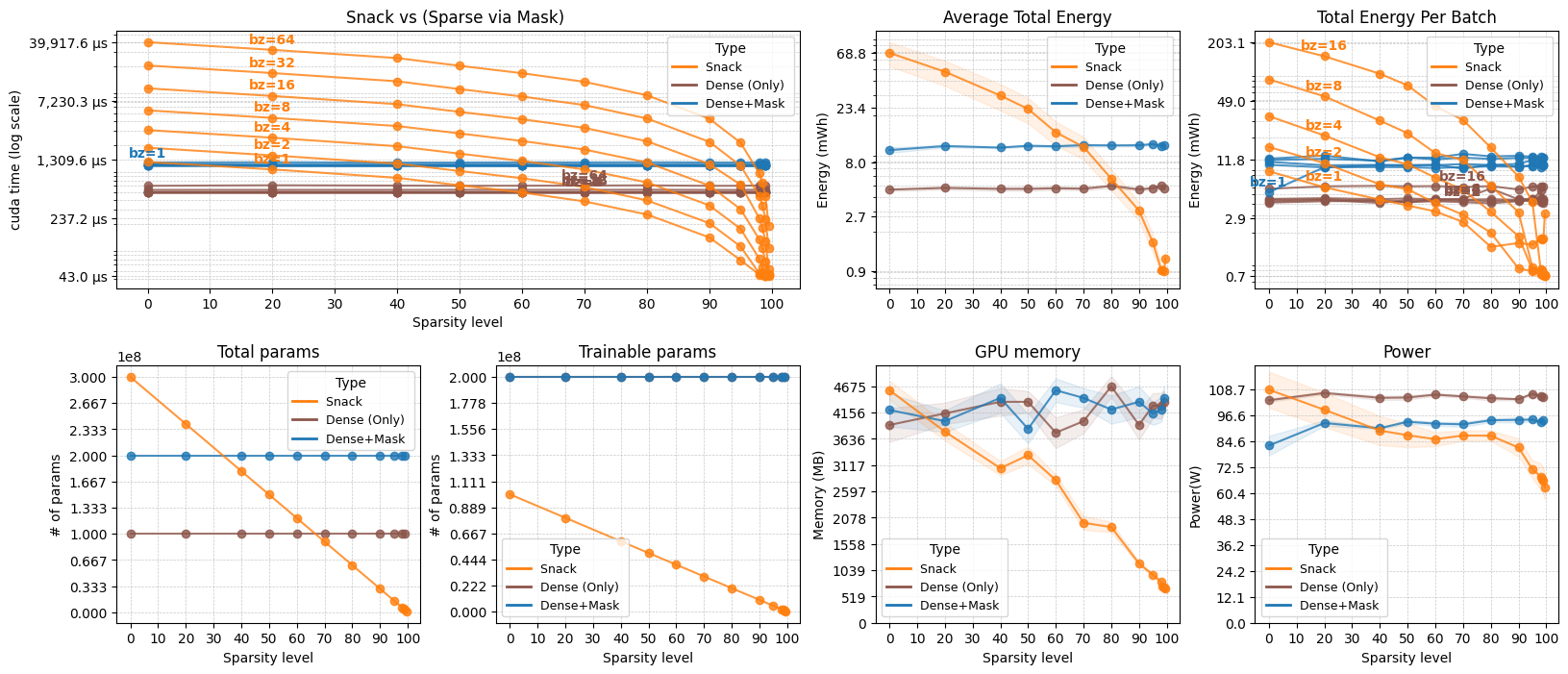}
    \caption{SNACK performance when SNACK layer dimension is 10000x10000}
    \label{fig:mask+dense-vs-snack-10000x10000-gpu-profile}
\end{figure}

\begin{figure}[H]
    \centering
    \includegraphics[width=\linewidth]{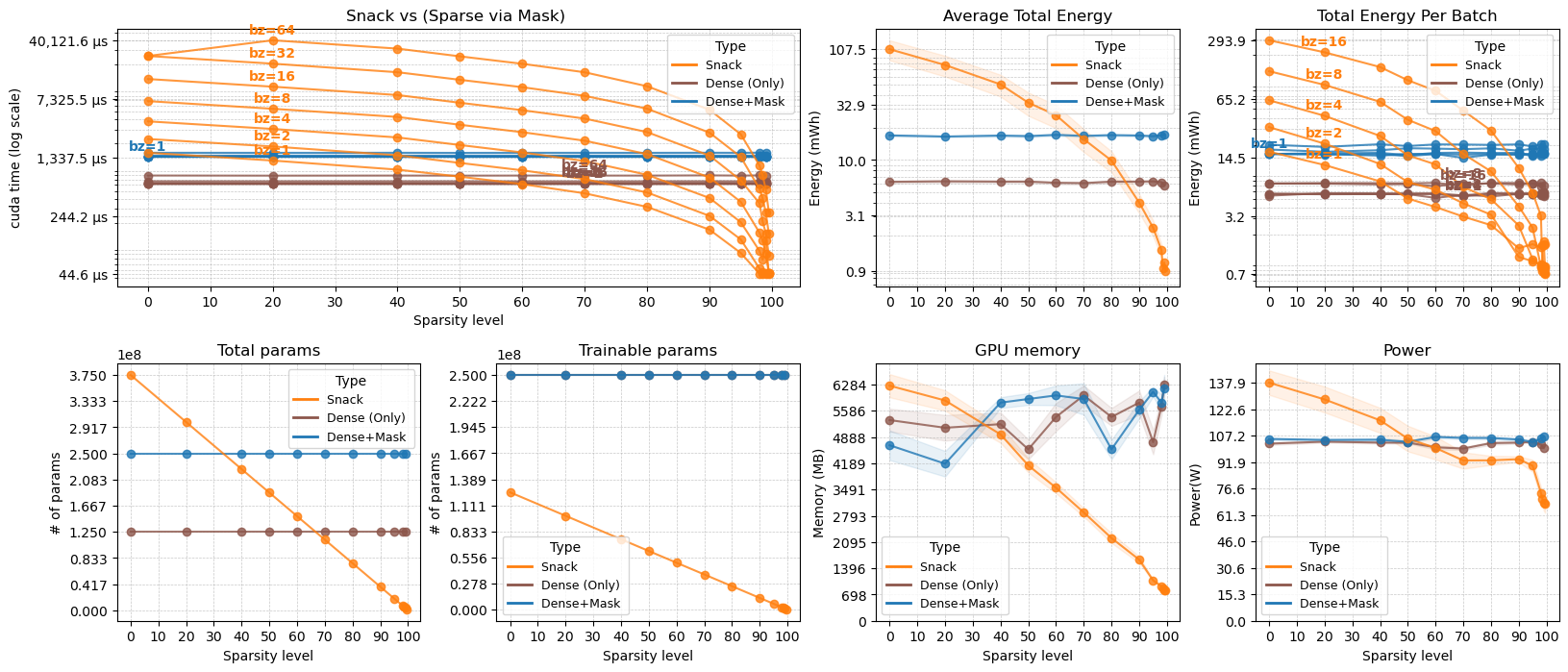}
    \caption{SNACK performance when SNACK layer dimension is 12500x10000}
    \label{fig:mask+dense-vs-snack-12500x10000-gpu-profile}
\end{figure}

In Figures~\ref{fig:mask+dense-vs-snack-2500x10000-gpu-profile}, \ref{fig:mask+dense-vs-snack-5000x10000-gpu-profile}, \ref{fig:mask+dense-vs-snack-7500x10000-gpu-profile}, \ref{fig:mask+dense-vs-snack-10000x10000-gpu-profile}, \ref{fig:mask+dense-vs-snack-12500x10000-gpu-profile}, we show more granular memory, speed, and energy results similar to Figure~\ref{fig:mask+dense-vs-snack-100-gpu-profile} in the main text. We see the findings that SNACK is faster, more memory optimized, and more energy efficient hold across various input sizes.

Our experiments showcase the efficiency of SNACK across various input size dimensions and a large hidden layer size (10000). Figures~\ref{fig:mask+dense-vs-snack-2500x10000-gpu-profile}--\ref{fig:mask+dense-vs-snack-12500x10000-gpu-profile} report on the comparison between SNACK, Dense+Mask, and Dense Only layers on energy consumption, speed, and memory utilization. SNACK achieves drastic improvements scaling with sparsity levels. For example in Table~\ref{tab:performance_comparison_snack_12500}, at extreme sparsity level ($\sigma =0.99 \text{ and input\_size = 12500})$ SNACK achieves execution times as low as $45.3$ $\mu$s (batch size 1), representing a \textbf{13.6$\times$ speedup} over Dense (Only) ($618.1$ $\mu$s) and a \textbf{30.0$\times$ speedup} over Dense+Mask ($1358.3$ $\mu$s). Energy consumption also decreases when we use SNACK, from $4.36$ mWh at 90\% sparsity (Dense+Mask) to $0.42$ mWh at 99\% sparsity using SNACK, representing \textbf{73\% less energy} than Dense (Only) and \textbf{90\% less energy} than Dense+Mask at the same sparsity level. Memory usage scales favorably, with SNACK using only $839$ MB at 99\% sparsity for $12500 \times 10000$ dimensions, representing \textbf{89\% memory reduction} compared to Dense+Mask ($7429$ MB).

\begin{table}[H]
\centering
\caption{Performance and environmental impact comparison of different implementations. All experiments use a 12500×10000 matrix with batch size of 1. Energy consumption (in mWh).}
\label{tab:performance_comparison_snack_12500}
\begin{tabular}{lcrrrrrr}
\toprule
\multirow{2}{*}{\textbf{Type}} & Sparsity & \textbf{Params} & \textbf{Memory} & \textbf{CUDA $\tau$} & \textbf{Power} & \textbf{Energy} & \textbf{CO$_2$} \\
 & \%& (k) & (MB) & ($\mu s$) & (W) & (mWh) & (mg) \\
\midrule
Dense+Mask    & 90 & 250,000 & 5995 & 1358.3 & 102.8 & 4.36 & 1.96 \\
SNACK (ours)  & 90 & \textbf{37,500} & \textbf{1863} & \textbf{163.1} & \textbf{100.2} & \textbf{0.39} & \textbf{0.18} \\
\midrule
Dense+Mask    & 95 & 250,000 & 6473 & 1357.4 & 98.9  & 4.20 & 1.89 \\
SNACK (ours)  & 95 & \textbf{18,750} & \textbf{1023} & \textbf{83.0}  & \textbf{111.5} & \textbf{0.43} & \textbf{0.19} \\
\midrule
Dense+Mask    & 99 & 250,000 & 7429 & 1358.3 & 103.3 & 4.38 & 1.97 \\
SNACK (ours)  & 99 & \textbf{3,750}  & \textbf{839}  & \textbf{45.3}  & \textbf{108.6} & \textbf{0.42} & \textbf{0.19} \\
\midrule
Dense (Only)  & -- & 125,000 & 6951 & 618.6  & 100.4 & 1.55 & 0.70 \\
\bottomrule
\end{tabular}
\end{table}

\subsection{Datacenter-class GPU sweep (NVIDIA A100)}\label{app:a100_spmm}

To test whether the SNACK-COO advantage observed on the consumer-class RTX~3070 (Table~\ref{tab:sparse_performance}) carries over to a datacenter-class accelerator, we re-ran the same SpMM micro-benchmark on an NVIDIA A100 across six leading-dimension sizes ($500$ through $10000$, square matrices, $95\%$ sparsity, batch~$1$, mean over $1{,}000$ multiplications). Table~\ref{tab:sparse_performance_a100} reports the resulting CUDA times. SNACK-COO remains the fastest sparse method at every size, with the constant-time \texttt{SpMM\_CSR(Cupy)} path closing to within $8$--$13\%$ at $\{8500, 10000\}$. At and above $5000{\times}5000$ SNACK-COO also beats the dense baseline outright, mirroring the RTX~3070 result. We exclude \texttt{SpMM\_BSR(Jax)} because the JAX/CUDA toolchain on the A100 host could not be brought up against the same library versions used in Table~\ref{tab:sparse_performance}; all other rows are matched to that table.

\begin{table}[H]
\footnotesize
\centering
\caption{\textbf{A100 replication of Table~\ref{tab:sparse_performance}.} CUDA time (ms) for sparse matrix multiplication implementations on an NVIDIA A100, sparsity $95\%$, batch size~$1$, square matrices, mean of $1{,}000$ multiplications. Best per-column value among sparse methods is \textbf{bold}; the \texttt{DenseMM} baseline is also \textbf{bold} where it remains the global best. $^{\dagger}$Simulates sparsity.}
\label{tab:sparse_performance_a100}
\begin{tabular}{lccccccc}
\toprule
\textbf{Method} & is sparse & \multicolumn{6}{c}{CUDA (ms)} \\
\cmidrule(lr){3-8}
& & 500 & 1000 & 5000 & 7500 & 8500 & 10000 \\
\midrule
SpMM\_CSR(Torch)             & \multirow{11}{*}{Yes} & 24.83 & 25.83 & 215.16 & 443.18 & 576.99 & 811.52 \\
SpMM\_CSR(Cupy)              &  & 11.35 & 11.05 & 11.63 & 11.45 & 14.47 & 15.68 \\
SpMM\_COO(Torch)             &  & 24.55 & 26.59 & 321.53 & 693.69 & 901.41 & 1260.55 \\
SNACK-COO (ours)             &  & \textbf{2.72} & \textbf{2.85} & \textbf{6.24} & \textbf{8.12} & \textbf{13.39} & \textbf{13.93} \\
SpMM\_COO(Cupy)              &  & 166.27 & 175.73 & 1153.05 & 2520.67 & 3410.45 & 4893.86 \\
SpMM\_CSR(cuSPARSE)          &  & 25.34 & 25.41 & 215.40 & 443.66 & 577.62 & 812.43 \\
SpMM\_COO(cuSPARSE)          &  & 25.03 & 27.84 & 305.23 & 700.28 & 890.00 & 1313.55 \\
SpMM\_CSR(Sputnik DL-opt)    &  & 3.05 & 3.07 & 8.10 & 14.12 & 17.57 & 24.01 \\
SpMM\_COO(Sputnik)           &  & 2.96 & 2.99 & 8.01 & 15.69 & 17.52 & 26.28 \\
FlashSparse                  &  & 27.56 & 37.08 & 329.91 & 706.73 & 835.93 & 1076.32 \\
DenseMM+Mask                 & $^{\dagger}$ & 4.38 & 4.44 & 46.68 & 100.90 & 130.43 & 174.30 \\
\midrule
DenseMM (baseline)           & no & \textbf{2.20} & \textbf{2.30} & 23.19 & 51.96 & 66.45 & 89.99 \\
\bottomrule
\end{tabular}
\end{table}

\paragraph{Cross-GPU consistency at batch~$1$.} The A100 numbers in Table~\ref{tab:sparse_performance_a100} preserve the qualitative ordering established on the RTX~3070 in Table~\ref{tab:sparse_performance}: SNACK-COO is the only sparse method that beats Dense+Mask across every size, the cuSPARSE / PyTorch / FlashSparse rows trail by an order of magnitude at small leading dimension, and the constant-time \texttt{SpMM\_CSR(Cupy)} path remains a strong outlier at the largest size only. At $5000{\times}5000$ SNACK-COO is $7.5\times$ faster than Dense+Mask on the A100 (vs.\ $7.0\times$ on the RTX~3070) and $3.7\times$ faster than the dense baseline (vs.\ $3.5\times$ on the RTX~3070), confirming that the batch-to-SM mapping (Section~\ref{subsec:coo_modified}) is not specific to consumer-class hardware.

\paragraph{Batch~$4$: the regime crossover predicted by Section~\ref{subsec:coo_modified}.} Table~\ref{tab:sparse_performance_a100_b4} repeats the A100 sweep at batch~$4$ over the same method set as Table~\ref{tab:sparse_performance_a100} (only \texttt{SpMM\_BSR(Jax)} is omitted, for the JAX/CUDA toolchain reason given above). SNACK-COO retains its lead at the two smallest sizes ($500$ and $1000$, where $n=4 \ll m$ still holds strongly) but is overtaken from $5000{\times}5000$ upward by Sputnik, with \texttt{SpMM\_COO(Sputnik)} winning at $\{5000, 8500\}$ and \texttt{SpMM\_CSR(Sputnik\,DL-opt)} winning at $\{7500, 10000\}$; the two Sputnik variants finish within ${\pm}10\%$ of each other at every size (mirroring their behaviour on the RTX~3070 in Table~\ref{tab:sparse_performance}), and the per-column-best at $10000{\times}10000$ is $21.44$\,ms versus SNACK-COO's $65.54$\,ms. This is exactly the crossover predicted by the parallel-efficiency analysis in Appendix~\ref{app:snack_details}: the SNACK-COO advantage is $\epsilon_p^{\text{SNACK-COO}} > \epsilon_p^{\text{standard}}$ for $n \ll m$, and as the batch grows or the matrix shrinks Sputnik's structured-sparse SpMM, which is tuned for moderate-batch / moderate-sparsity, becomes the better choice. This is the operational reason the SNACK \emph{layer} keeps the SpMM backend swappable (Section~\ref{subsec:impl}) and why Appendix~\ref{app:gamlp_setup} trains GAMLP through Sputnik but serves it through SNACK-COO. Note that even at this less-favorable batch size, SNACK-COO still beats \texttt{Dense+Mask} on every size and beats the dense baseline at sizes $\geq 5000$, so the kernel remains a valid choice whenever Sputnik is unavailable.

\begin{table}[H]
\footnotesize
\centering
\caption{\textbf{A100 SpMM benchmark at batch size~$4$.} CUDA time (ms) for sparse matrix multiplication implementations on an NVIDIA A100, sparsity $95\%$, batch size~$4$, square matrices, mean of $1{,}000$ multiplications. Same method set as Table~\ref{tab:sparse_performance_a100}; \texttt{SpMM\_BSR(Jax)} is omitted because the JAX/CUDA toolchain on the A100 host could not be brought up at the matched library versions. Best per-column value among sparse methods is \textbf{bold}; the \texttt{DenseMM} baseline is also \textbf{bold} where it remains the global best. $^{\dagger}$Simulates sparsity.}
\label{tab:sparse_performance_a100_b4}
\begin{tabular}{lccccccc}
\toprule
\textbf{Method} & is sparse & \multicolumn{6}{c}{CUDA (ms)} \\
\cmidrule(lr){3-8}
& & 500 & 1000 & 5000 & 7500 & 8500 & 10000 \\
\midrule
SpMM\_CSR(Torch)             & \multirow{10}{*}{Yes} & 27.46 & 28.25 & 248.35 & 528.04 & 675.79 & 950.01 \\
SpMM\_CSR(Cupy)              &  & 12.15 & 12.19 & 12.81 & 21.32 & 27.07 & 37.01 \\
SpMM\_COO(Torch)             &  & 27.37 & 30.03 & 342.17 & 767.56 & 979.77 & 1392.60 \\
SNACK-COO (ours)             &  & \textbf{2.98} & \textbf{3.05} & 21.12 & 36.90 & 52.17 & 65.54 \\
SpMM\_COO(Cupy)              &  & 180.59 & 185.85 & 1267.72 & 2800.83 & 3618.27 & 5160.39 \\
SpMM\_CSR(cuSPARSE)          &  & 27.49 & 28.28 & 248.62 & 528.62 & 676.53 & 951.06 \\
SpMM\_COO(cuSPARSE)          &  & 27.27 & 29.93 & 341.93 & 767.45 & 981.85 & 1391.79 \\
SpMM\_CSR(Sputnik DL-opt)    &  & 3.36 & 3.33 & 8.24 & \textbf{12.29} & 16.08 & \textbf{21.44} \\
SpMM\_COO(Sputnik)           &  & 3.26 & 3.24 & \textbf{8.15} & 13.66 & \textbf{16.03} & 23.46 \\
FlashSparse                  &  & 30.04 & 39.86 & 369.50 & 774.58 & 922.03 & 1140.90 \\
DenseMM+Mask                 & $^{\dagger}$ & 4.66 & 4.68 & 50.20 & 110.71 & 139.38 & 194.47 \\
\midrule
DenseMM (baseline)           & no & \textbf{2.34} & \textbf{2.35} & 25.23 & 55.63 & 70.04 & 97.72 \\
\bottomrule
\end{tabular}
\end{table}

\section{Extended Related Works}
\label{app:extended_related_works}
This appendix expands the discussion in Section~\ref{sec:related_works} with material that we omit from the main text for brevity.

\paragraph{Early pruning, lottery tickets, and pruning-at-init.} Early sparsity work focused on post-training pruning, notably Optimal Brain Damage (OBD)~\cite{NIPS1989_6c9882bb} and Optimal Brain Surgeon~\cite{hassibi1993optimal}, which removed weights with minimal impact on performance. The Lottery Ticket Hypothesis~\cite{frankle2018lottery} showed that dense models contain sparse sub-networks capable of learning independently when trained from their original initialization. A complementary line removes weights \emph{before} training: SNIP~\cite{lee2019snip} prunes by single-shot connection sensitivity, and SynFlow~\cite{tanaka2020pruning} iteratively conserves synaptic flow without using any data; both establish a sparse mask once and then train dense-on-mask, in contrast to the prune-and-regrow loop of DST. Modern pruning techniques~\cite{frantar2023sparsegpt, zhu2017prune, liang2021pruning} compress models for resource-constrained deployment and typically apply binary masks over dense weights to simulate sparsity~\cite{zhu2017prune}. SparseGPT~\cite{frantar2023sparsegpt} enables one-shot unstructured pruning of large language models with minimal performance degradation, without retraining. Hardware-friendly transformations such as SPDF~\cite{thangarasa2023spdf} combine unstructured sparse pre-training with dense fine-tuning, achieving a $2.5\times$ FLOP reduction. Sparse-IFT~\cite{thangarasa2023sparse} proposes drop-in transformations that preserve dense FLOP counts while improving accuracy, but require specialized hardware kernels. The structured/unstructured distinction~\cite{wen2016learning} is central: structured patterns are regular and hardware-friendly; unstructured sparsity is more flexible but harder to accelerate.

\paragraph{Dynamic sparse training history.} Beyond SET~\cite{mocanu2018scalable} and RigL~\cite{evci2020rigging}, two earlier works are central to the DST lineage that SNACK plugs into. Deep rewiring~\cite{bellec2018deep} formalised topology evolution as a sampling process governed by a posterior over connectivities and showed that very sparse networks could be trained from scratch on small benchmarks. Sparse-from-Scratch~\cite{dettmers2019sparse} introduced momentum-based regrowth and demonstrated, on the available hardware of the time, that DST can match dense baselines without any dense pretraining stage. SNACK is agnostic to the specific prune-and-regrow rule: any of SET, RigL, deep rewiring, or sparse-from-scratch can drive its uniform connection initializer.

\paragraph{Memory-efficient LLM training.} Two lines of work attack the LLM training memory wall along axes orthogonal to a truly sparse weight tensor. \cite{yuan2021mest} (MEST) keeps the weight tensor dense at compute time but stores optimizer states only for active parameters, plus a soft memory-economic schedule on the regrowth set; the storage savings stack with SNACK's, since SNACK additionally removes the dense weight and gradient tensors themselves. \cite{zhao2024galore} (GaLore) projects gradients onto a low-rank subspace before applying Adam, reducing optimizer-state memory at the cost of a periodic SVD; this is orthogonal to SNACK because the savings come from the gradient/optimizer side rather than the weight side, and the two could be combined (a sparse weight tensor with low-rank optimizer states) without architectural conflict. At LLM-pretraining scale, \cite{agarwalla2024enabling} pretrains high-sparsity Llama variants and pairs them with a tuned sparse-deployment stack; their kernel choice (CSR with structured 2:4 fallbacks) targets the moderate-batch deployment regime where Sputnik dominates, whereas SNACK-COO targets the small-batch regime they leave for further work.

\paragraph{Sparse formats and earlier GPU SpMM kernels.} The CSR format dominates GPU-based SpMM implementations and leverages warp-level threads to distribute row-wise computation~\cite{bell2009implementing}. Block Sparse Row (BSR) extends CSR by grouping non-zeros into $n \times n$ blocks: it behaves similarly to CSR for small block sizes and trades memory for cache efficiency at larger block sizes~\cite{merrill2016merge}. Customized kernel fusion and load-balancing techniques further improved SpMM on GPUs~\cite{yang2018design}. Energy efficiency in SpMM/SpMV is closely tied to reducing computational load: the less time required per multiplication, the less energy is required to power the GPU/TPU/CPU.

\paragraph{Block-sparse Transformer pretraining (BLaST).} BLaST~\cite{okanovic2025blast} couples a fused Triton block-SpMM kernel (BCSC layout, Tensor-Core MMA-aligned $\mathrm{blk\_M}\!\times\!\mathrm{blk\_N}$ tiles) with a block prune-and-grow training loop and a second-order \emph{oLLM} pruner inspired by Optimal Brain Surgeon~\cite{hassibi1993optimal}. It reaches up to $95\%$ MLP sparsity at a median $\le 2.25\%$ accuracy drop and reports a $2.2\times$ end-to-end inference speedup for Llama-3.2 across $16$ GH200 GPUs, with up to $4.45\times$ inference-memory reduction. BLaST is therefore complementary to SNACK on two axes. First, it requires \emph{structured} (block) sparsity: the kernel's speedup hinges on the BCSC block size matching MMA fragment shapes, whereas SNACK targets the unstructured patterns produced by SET/RigL-style DST. Second, BLaST is not truly sparse end-to-end: during training the dense weight and gradient tensors are kept intact and only the pruned forward/backward path runs through BCSC, so the dense-tensor materialization that SNACK explicitly avoids is still paid at training time (and amortized only as $s_i$ grows). The two systems thus occupy adjacent regimes -- BLaST for block-sparse, multi-GPU LLM pretraining at moderate batch sizes, SNACK for unstructured, small-batch, high-sparsity training and inference on commodity GPUs.

\paragraph{Truly sparse implementations on CPU and IPU.} Using sparse data structures in SciPy and Cython,~\cite{pmlr-v139-liu21y} implemented and trained with DST a truly sparse Multi-Layer Perceptron (SET-MLP) with one million neurons on a single CPU core, with single-core execution being the practical limitation of their approach.~\cite{curci2021truly} parallelized this approach across multiple CPU cores but reported a hard dependence on CPU.~\cite{Atashgahi2022} adjusted the approach to autoencoders and showed that a truly sparse implementation on one CPU core can outperform the dense GPU baseline in run-time and energy efficiency, owing to DST's quadratic parameter reduction. PopSparse~\cite{li2023popsparse} leverages IPU hardware to deliver $90\%$-sparsity speedups. SNACK is, in contrast, the first such truly sparse PyTorch building block on standard GPUs, with no hardware-specific sparsity support required.

\paragraph{Sparse accelerators and adjacent kernels.} Several theoretical sparse accelerators have been proposed, including SpArch's fixed outer-product dataflow and Flexagon's reconfigurable multi-dataflow architecture~\cite{9065428, munoz2023flexagon}. Gamma~\cite{zhang2021gamma} leverages Gustavson's algorithm with specialized fiber cache and merger hardware to accelerate sparse matrix multiplication. \cite{schultheis2023towards} demonstrate custom CUDA kernels for uniform sparse matrix multiplication in extreme classification, where each output neuron receives the same number of inputs, enabling memory-efficient training with 500K labels on commodity GPUs. These directions are complementary to SNACK, which targets unstructured sparsity on commodity hardware without architectural changes.

\section{SNACK Algorithmic Details}\label{app:snack_details}
This appendix is the technical companion to Section~\ref{sec:our_method}: a parallel-efficiency proof for SNACK-COO (referenced from Section~\ref{subsec:coo_modified}); the geometric-distribution analysis of the uniform initializer's expected cost (referenced from Algorithm~\ref{alg:uni_init}); the full chain-rule derivation of SNACK's sparse forward and backward, including the mini-batch generalization (referenced from Section~\ref{subsec:sparse_grad}); the CUDA/\texttt{autograd} implementation plumbing (referenced from Section~\ref{subsec:impl}); and a closing summary of how end-to-end true sparsity is preserved across the entire pipeline.

\paragraph{Parallel-efficiency analysis for SNACK-COO (Section~\ref{subsec:coo_modified}).}
The standard COO scheme creates $m$ parallel row tasks with the batch dimension as an outer loop. When the layer is much wider than the batch, $m > |\mathrm{SMs}| \cdot \mathrm{warps}_{\mathrm{SM}}$, many SMs sit idle. SNACK-COO replaces this with a 2D block grid $\mathcal{G} = \{(N_b, n)\}$ where $N_b = \lceil nnz / T \rceil$, so the task pool is $N_b \cdot n$ blocks (i.e., $nnz \cdot n$ thread tasks). The work distribution per SM becomes
\[
\frac{N_b \cdot n}{|\mathrm{SMs}|} = \frac{\lceil nnz / T \rceil \cdot n}{|\mathrm{SMs}|},
\]
which scales with the non-zero workload rather than with the matrix height $m$. Within a block, thread $\tau$ in block $(i, b)$ reads the COO triple at index $k = i T + \tau$ ($k < nnz$) and atomically updates $C_{A_r[k],b} \mathrel{+}= A_v[k] \cdot B_{A_c[k],b}$. The time complexity remains $\mathcal{O}(nnz \cdot n)$, but the parallel efficiency
\[
\epsilon_p = \frac{T_{\mathrm{seq}}}{|\mathrm{SMs}| \cdot T_{\mathrm{par}}}, \qquad \epsilon_p^{\text{SNACK-COO}} > \epsilon_p^{\text{standard}}
\]
improves because $T_{\mathrm{par}}^{\text{SNACK-COO}} < T_{\mathrm{par}}^{\text{standard}}$ at small $n$. Atomic adds are required when multiple non-zeros write to the same output row, but the batch-wise block organization confines contention to one output column per block, reducing it relative to the row-wise scheme. Threads within a block process consecutive non-zeros for the same column $b$, which also improves cache locality on $B$.

\paragraph{Uniform Initializer cost (Algorithm~\ref{alg:uni_init}).}
Algorithm~\ref{alg:uni_init} draws $nnz = (1-\eta)\,mk$ unique pairs by rejection: at step $i$, the probability of drawing a fresh pair is $p_i = 1 - i/(mk)$, giving a geometric expected number of trials of $\mathbb{E}[1/p_i] = mk/(mk - i)$. Summing over the $nnz$ steps,
\[
\mathbb{E}[\text{trials}] = \sum_{i=0}^{(1-\eta)mk} \frac{mk}{mk - i} \;\approx\; \int_{0}^{(1-\eta)mk} \frac{mk}{mk - i}\,di = -mk \ln \eta, \qquad \eta \in (0, 1),
\]
where the harmonic approximation $\sum_{i=a}^{b} 1/i \approx \ln(b/a)$ is used. The initializer is therefore essentially linear in $nnz$, and the rejection penalty $\zeta = \mathbb{E}[\text{trials}] / nnz = -\ln \eta / (1-\eta)$ is bounded: $\zeta \to 1$ as $\eta \to 1$ (very sparse, almost no collisions), and $\zeta$ grows only logarithmically as $\eta$ shrinks toward the dense regime.

\paragraph{Full gradient derivation (Section~\ref{subsec:sparse_grad}).}
For a SNACK layer with sparse $W \in \mathbb{R}^{m \times n}$, input $X \in \mathbb{R}^{n \times b}$, and pre-activation $A = WX \in \mathbb{R}^{m \times b}$, the chain rule gives
\[
\frac{\partial \mathcal{L}}{\partial W} = \frac{\partial \mathcal{L}}{\partial A}\,\frac{\partial A}{\partial W} = \nabla A \cdot X^\top, \qquad \frac{\partial \mathcal{L}}{\partial X} = \Bigl(\frac{\partial A}{\partial X}\Bigr)^{\!\!\top}\!\!\frac{\partial \mathcal{L}}{\partial A} = W^\top\,\nabla A,
\]
where $\nabla A = \partial \mathcal{L}/\partial A \in \mathbb{R}^{m \times b}$. The input-gradient term $W^\top \nabla A$ is sparse-times-dense and reuses SNACK-COO directly: $W^\top$ inherits $W$'s sparsity by swapping row and column index arrays, introducing no new non-zeros. Expanding the weight-gradient product entry-wise,
\[
\Bigl(\frac{\partial \mathcal{L}}{\partial W}\Bigr)_{i,j} = \sum_{t=1}^{b} \nabla A_{i,t}\,X_{j,t},
\]
which is a sum of $b$ rank-1 outer products: $\partial \mathcal{L}/\partial W = \sum_{t=1}^{b} \nabla A^{(t)} (X^{(t)})^\top$. A dense backward evaluates all $mn$ entries of this matrix; SNACK evaluates only $(i,j) \in \mathrm{supp}(W)$, since entries outside the support are zeroed by the sparsity mask and contribute nothing to the SGD update. This drops the per-step cost from $\mathcal{O}(mn b)$ to $\mathcal{O}(nnz \cdot b)$. Activation functions act element-wise on $A$ and are unaffected by $W$'s sparsity~\cite{hagan1997neural}.

\paragraph{Implementation plumbing (Section~\ref{subsec:impl}).}
The SNACK-COO kernel parallelizes over sparse-matrix cells rather than over batch entries, which saturates the SMs from the non-zero workload alone and is what makes the kernel favorable at small $n$. Within a block we issue \texttt{\_\_syncthreads()} before atomic writes to avoid races, and we use $T = 256$ threads per block as a default that balances occupancy against scheduling overhead. The CUDA launch is wrapped in a C++/Torch extension that converts the COO triple from PyTorch tensors to flat \texttt{float}/\texttt{int} arrays and back. The sparse outer-product gradient is a separate one-dimensional CUDA kernel that loops over the $nnz$ entries of $\mathrm{supp}(W)$ and writes back the accumulated $\sum_t \nabla A_{i,t} X_{j,t}$ for each entry. The PyTorch \texttt{nn.Module} wraps both kernels behind a custom \texttt{torch.autograd.Function}, accepts a pluggable \texttt{Initializer} (default: the C++ uniform initializer of Algorithm~\ref{alg:uni_init}), and stores weights as the COO triple $(\mathbf{A}_v, \mathbf{A}_r, \mathbf{A}_c)$ on-device for the entire training run.

\paragraph{End-to-end sparsity.}
Across the full pipeline, SNACK preserves true sparsity at every step: (i) the forward pass runs SpMM (SNACK-COO by default, or any pluggable backend) directly on the COO triple and never expands $W$ to dense; (ii) the backward pass uses the sparse outer product over $\mathrm{supp}(W)$ and never instantiates the $m \times n$ weight-gradient matrix; (iii) DST topology updates use the uniform initializer of Algorithm~\ref{alg:uni_init}, which draws non-zeros directly with no dense $m \times k$ probability tensor. Consequently no step of training or inference allocates a dense $m \times n$ weight or gradient tensor, which is the property the memory and energy measurements of Sections~\ref{sec:exps}--\ref{sec:end_to_end_experiments} exploit.

\section{GAMLP Graph-Workload Training Setup}\label{app:gamlp_setup}
This appendix details the training recipe behind the GAMLP-shaped sparse-inference results of Table~\ref{tab:spmm_inference_compact} (Section~\ref{sec:end_to_end_experiments}). Inference is reported at batch~$=1$ on the trained sparse weights; training itself is run at the natural mini-batch size for graph workloads, $b=4096$, where the leading dimension is no longer ``small'' relative to the layer width and SNACK-COO's batch-to-SM mapping ceases to be the right fit. We therefore train through the SNACK layer with a \emph{Sputnik} SpMM backend, which dominates at moderate sparsity and large batches~\cite{gale2020sparse}, and only swap to SNACK-COO at inference time. This is exactly the backend-agnostic story of Section~\ref{sec:our_method}: the SNACK layer is fixed (sparse outer-product gradient, uniform initializer, COO weights), and the SpMM backend is chosen per regime.

\paragraph{Model and data.} We use \texttt{R\_GAMLP\_RLU}~\cite{zhang2022graph} on the OGB \texttt{ogbn-products} node-classification benchmark~\cite{hu2020open}, with hidden dimension $512$, $5$ feature-propagation hops, and $2/2/2$ layers in the three GAMLP stages (default \texttt{R\_GAMLP\_RLU} configuration). Pre-processing, residual connections, and batch norm are enabled; dropout, attention dropout, and label dropout are $0.5$, with $\alpha=0.5$ for the residual mix. The sparsifiable substrate is the $5000{\times}5000$ feature-head projection used at inference; embeddings, label propagation, and the classification head are kept dense.

\paragraph{Optimization.} Training uses the Adam optimizer at learning rate $10^{-3}$ with cross-entropy loss for $40$ epochs at batch size $4096$ on a single GPU. We use the same dense GAMLP hyperparameters as the public reference and only swap the feature-head projection from a dense \texttt{nn.Linear} to a SNACK module with the Sputnik backend; everything else (input pipeline, scheduler, weight decay, seed) follows the upstream GAMLP defaults. Dense and \texttt{Dense+Mask} baselines for the same backbone are trained under an identical recipe, the only difference being the feature-head implementation.

\paragraph{Sparsity schedule and DST.} Each run is initialized at its target sparsity $\eta\in\{0.9, 0.95, 0.99\}$ and trained with a magnitude-based prune-and-grow update: every $5$ epochs (until epoch $25$), we prune the $\zeta = 5\%$ smallest-magnitude non-zeros and regrow an equal number of unique $(i,j)$ coordinates with the uniform connection initializer of Algorithm~\ref{alg:uni_init}. After epoch $25$ the topology is frozen and only weight values continue to update, which gives the optimizer a stable support to converge on. The inference results in Table~\ref{tab:spmm_inference_compact} use the resulting frozen SNACK checkpoints, loaded directly into the SNACK-COO inference path with no re-pruning.

\paragraph{Why Sputnik for training and SNACK-COO for inference.} SNACK-COO's parallel-efficiency analysis (Appendix~\ref{app:snack_details}) shows the kernel's advantage relies on $n \ll m$, i.e.\ batch much smaller than the layer width. At training-time batch $4096$ that condition is reversed and Sputnik's structured-sparse SpMM is the better fit; at inference-time batch~$1$ the condition is restored and SNACK-COO is the better fit. Both paths use \emph{the same} SNACK layer, the same COO weights, and the same outer-product gradient; only the SpMM backend changes. This is the concrete realization of the backend-agnostic claim made in Section~\ref{subsec:impl} and explains why the same module can deliver Table~\ref{tab:gpt2_training_compact}'s training memory savings under Sputnik and Table~\ref{tab:spmm_inference_compact}'s inference speedups under SNACK-COO. The full training script and reproducibility commands are available at \url{https://github.com/JafarBadour/Snack-SNN} under \texttt{experiments/gamlp\_snack\_benchmark/}.

\begin{table*}[t]
    \centering
    \footnotesize
    \caption{\textbf{GAMLP-style dynamic sparse training on \texttt{ogbn-products}} at sparsity $\eta=0.9$, batch size~$4096$, $40$ epochs on a single GPU. We report best and final \emph{validation} accuracy, mean per-step training latency (ms), per-step energy (mJ), and peak GPU allocation (MB). The \texttt{SNACK-COO} (ours) row reflects training executed through the SNACK layer with the SNACK-COO backend, included to expose the $n \ll m$ regime mismatch at $b=4096$ predicted in Appendix~\ref{app:snack_details}; under the recommended per-regime backend choice the SNACK layer is trained with the Sputnik backend and only swapped to SNACK-COO at inference time (Table~\ref{tab:spmm_inference_compact}). Best per-column value is in \textbf{bold}.}
    \label{tab:gamlp_training_compact}
    \begin{tabular}{lrrrrr}
    \toprule
    Method & Best val acc & Final val acc & Latency\_step\_ms & Energy\_step\_mJ & Peak Mem (MB) \\
    \midrule
    Dense              & \textbf{0.9232} & \textbf{0.9191} & \textbf{17.11} & 1700.3 & 6823 \\
    Dense+Mask         & 0.9174 & 0.9140 & 17.34 & \textbf{1694.3} & 6840 \\
    SNACK-COO (ours)   & 0.9172 & 0.9101 & 160.39 & 17177.1 & \textbf{6784} \\
    \bottomrule
    \end{tabular}
\end{table*}

\paragraph{Empirical validation of the per-regime backend choice (Table~\ref{tab:gamlp_training_compact}).} The GAMLP training measurements directly back the argument above: at the natural graph-training batch $4096$, the layer-width-to-batch ratio violates SNACK-COO's $n \ll m$ assumption, and SNACK-COO accordingly pays a $\approx 9.4\times$ per-step latency tax ($160.39$ vs.\ $17.11$\,ms) and a $\approx 10\times$ per-step energy tax ($17{,}177$ vs.\ $1{,}700$\,mJ) over \texttt{Dense}. This is precisely the regime the paragraph above flags as Sputnik's territory, which is why the recommended per-regime configuration trains the SNACK layer through the Sputnik backend at $b=4096$ (whose moderate-sparsity, large-batch dominance is documented in~\cite{gale2020sparse} and reproduced in our A100 sweeps in Appendix~\ref{app:a100_spmm}, where the Sputnik variants overtake SNACK-COO from $5000{\times}5000$ upward at batch~$4$) and only swaps to the SNACK-COO backend at the inference-time batch~$1$ regime, which is the configuration whose memory and latency we report in Table~\ref{tab:spmm_inference_compact}. Peak training memory is already minimally smaller for SNACK-COO ($6784$ vs.\ $6823$\,MB for \texttt{Dense}); the modest gap is consistent with the architectural ceiling noted in Section~\ref{sec:end_to_end_experiments}: only the $5000{\times}5000$ feature-head projection is sparsified while the rest of the GAMLP backbone (multi-hop feature propagation buffers, embeddings, classifier head) remains dense. Validation accuracy is essentially matched: \texttt{SNACK-COO} reaches $0.9101$ final / $0.9172$ best vs.\ \texttt{Dense+Mask}'s $0.9140$ final / $0.9174$ best, a sub-half-point gap that is independent of the SpMM backend, since both \texttt{SNACK-COO} and \texttt{SNACK(Sputnik)} share the same uniform sparse initializer, sparse outer-product gradient, and magnitude prune-and-grow schedule, so any kernel-level swap is mathematically a reordering and will not move accuracy materially. The remaining accuracy headroom is a DST-recipe question (initialization, prune-and-grow schedule, dropout/regularization), orthogonal to the kernel contribution of this paper.
\clearpage

\end{document}